\pdfoutput=1
\documentclass[11pt]{article}

\usepackage[final]{acl}

\usepackage{times}
\usepackage{latexsym}
\usepackage{pifont}
\usepackage{amsmath}
\usepackage{bbm}

\usepackage[T1]{fontenc}
\usepackage[utf8]{inputenc}

\usepackage{microtype}

\usepackage{inconsolata}

\usepackage{graphicx}
\usepackage{amssymb}
\usepackage{booktabs,multirow,longtable,makecell}
\usepackage{setspace}

\title{Themis: A Reference-free NLG Evaluation Language Model with Flexibility and Interpretability}

\author{Xinyu Hu,\: Li Lin,\: Mingqi Gao,\: Xunjian Yin,\: Xiaojun Wan\\
Wangxuan Institute of Computer Technology, Peking University\\
\{huxinyu, gaomingqi, xjyin, wanxiaojun\}@pku.edu.cn \\
efsotr\_l@stu.pku.edu.cn}

\begin{document}
\maketitle
\begin{abstract}
The evaluation of natural language generation (NLG) tasks is a significant and longstanding research area. With the recent emergence of powerful large language models (LLMs), some studies have turned to LLM-based automatic evaluation methods, which demonstrate great potential to become a new evaluation paradigm following traditional string-based and model-based metrics. However, despite the improved performance of existing methods, they still possess some deficiencies, such as dependency on references and limited evaluation flexibility. Therefore, in this paper, we meticulously construct a large-scale NLG evaluation corpus \textbf{NLG-Eval} with annotations from both human and GPT-4 to alleviate the lack of relevant data in this field. Furthermore, we propose \textbf{Themis}, an LLM dedicated to NLG evaluation, which has been trained with our designed multi-perspective consistency verification and rating-oriented preference alignment methods. Themis can conduct flexible and interpretable evaluations without references, and it exhibits superior evaluation performance on various NLG tasks, simultaneously generalizing well to unseen tasks and surpassing other evaluation models, including GPT-4.

\end{abstract}

\section{Introduction}

Automated evaluation is crucial for natural language generation tasks, as it measures the performance of related models and consequently promotes the development of NLG research. In the early years, traditional string-based evaluation metrics, such as BLEU~\citep{DBLP:conf/acl/PapineniRWZ02}, were commonly used. Despite their convenience, surface-level matching cannot reliably evaluate texts as they are easily affected by perturbations~\citep{DBLP:conf/acl/HeZ0KCGT23}, and previous work~\citep{DBLP:conf/emnlp/SulemAR18} has indicated their low correlation with human evaluations. With the development of pre-trained language models and related corpora, pre-trained model-based evaluation metrics such as BARTScore~\citep{DBLP:conf/nips/YuanNL21} and COMET~\citep{DBLP:conf/emnlp/ReiSFL20} have then been proposed. But their performance remains unsatisfactory compared with human evaluation, and they cannot conduct evaluations on customized aspects, like coherence.

\begin{figure}[t]
\centering
\includegraphics[width=0.482\textwidth]{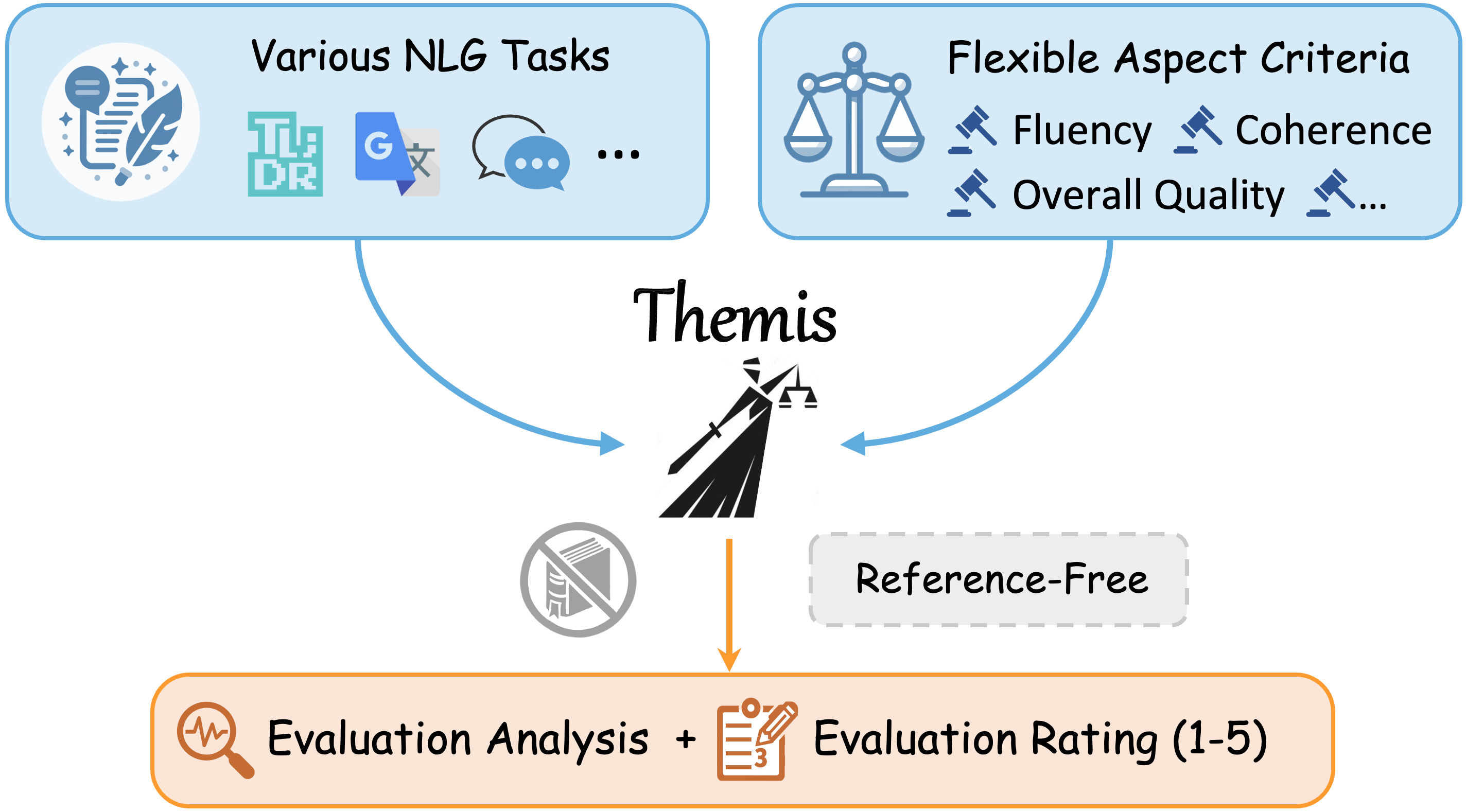}
\caption{\textbf{Themis} is capable of evaluating various NLG tasks based on flexible evaluation aspects and criteria without references and providing corresponding analyses along with the evaluation ratings.}
\label{fig:1}
\end{figure}

Recently, large language models (LLMs) such as ChatGPT and LLaMA~\citep{DBLP:journals/corr/abs-2302-13971} have emerged and demonstrated unprecedented performance in instruction following and open-domain generation, which shows great potential for LLM-based automated evaluation to become a new paradigm~\citep{DBLP:journals/corr/abs-2303-04048, DBLP:journals/corr/abs-2402-01383}. Existing related studies can be divided into two major categories: directly prompting LLMs with different optimization methods~\citep{DBLP:conf/emnlp/LiuIXWXZ23, DBLP:conf/emnlp/ChiangL23, DBLP:conf/coling/LiuYHZHWDSZ24, DBLP:conf/eamt/KocmiF23} and fine-tuning LLMs with annotated data~\citep{DBLP:journals/corr/abs-2310-05470, DBLP:journals/corr/abs-2310-00752, DBLP:conf/emnlp/XuWPSFWL23, DBLP:journals/corr/abs-2310-08491}. However, these approaches still have some limitations: the first category often relies on proprietary LLMs such as GPT-4, which are high-cost and possibly irreproducible, while the second category tends to have weaknesses of dependence on references, lacking flexibility, or poor interpretability.

In this paper, we propose \textbf{Themis}, an 8B-parameter LLM specifically designed and trained for NLG evaluation with more comprehensive capabilities. Our Themis can evaluate various NLG tasks, including uncommon ones like question-answering evaluation (\textbf{Versatility}), in a reference-free manner (\textbf{Independence}). Moreover, it allows for specific and customized evaluation aspects and criteria, including overall quality and more fine-grained aspects (\textbf{Flexibility}), and its evaluation contains corresponding analysis and explanation together with the rating (\textbf{Interpretability}), as shown in Figure~\ref{fig:1}. We believe that an ideal evaluator should be convenient to use and possess these characteristics. The comparison between related methods and Themis is shown in Table~\ref{tab:1}.

To obtain high-quality training data for NLG evaluation, we conducted a comprehensive survey of relevant studies and collected corresponding resources, finally selecting 58 evaluation datasets with human annotations across 9 common NLG tasks. Given the importance of evaluation criteria, we meticulously proofread each dataset and manually supplemented the missing descriptions. In addition, GPT-4 is treated as an additional annotator to supplement and validate the collected data while providing evaluation analyses. Ultimately, we constructed a large-scale evaluation corpus, NLG-Eval, which contains about 0.5 million samples with meta information. It aims to alleviate the issue of scattered and scarce data in the area of NLG evaluation and facilitate relevant research.

Furthermore, we proposed a multi-perspective consistency verification method to select relatively more reliable data from the constructed NLG-Eval corpus. Empirical methods are also employed to ensure sufficient diversity and balanced distribution of the data as much as possible, resulting in approximately 67K training samples. Moreover, we designed specific preference alignment, guiding the construction and utilization of preference data through evaluation ratings, to improve the evaluation capabilities of the fine-tuned model. Experimental results show that our Themis achieves better overall evaluation performance over previous evaluation models on common NLG tasks, including summarization, story generation, and so on. We also conducted more in-depth analyses, including generalization tests on unseen tasks like the instruction-following evaluation as well as aspect-targeted perturbation tests to verify the reliability.

\begin{table}
\centering
\small
\setlength{\tabcolsep}{4pt}
\renewcommand{\arraystretch}{1.25}
\begin{tabular}{cccccc}
\toprule
\textbf{Method} & \textbf{Vers.} & \textbf{Inde.} & \textbf{Flex.} & \textbf{Inte.} & \textbf{Open.} \\
\midrule
UniEval & \ding{55} & \ding{55} & \ding{51} & \ding{55} & \ding{51} \\
G-Eval & \ding{51} & \ding{51} & \ding{51} & \ding{51} & \ding{55} \\
X-Eval & \ding{51} & \ding{55} &  \ding{51} & \ding{55} & \ding{55} \\
Prometheus & \ding{51} & \ding{55} & \ding{51} & \ding{51} & \ding{51} \\
Auto-J & \ding{51} & \ding{51} & \ding{55} & \ding{51} & \ding{51} \\
InstructScore &\ding{51} & \ding{55} & \ding{55} & \ding{51} & \ding{51} \\
TIGERScore & \ding{51} & \ding{51} & \ding{55} & \ding{51} & \ding{51} \\
\textbf{Themis (Ours)} & \ding{51} & \ding{51} & \ding{51} & \ding{51} & \ding{51} \\
\bottomrule
\end{tabular}

\caption{Comparisons of our Themis with currently common evaluation models, including UniEval~\citep{DBLP:conf/emnlp/Zhong0YMJLZJH22}, G-Eval~\citep{DBLP:conf/emnlp/LiuIXWXZ23}, X-Eval~\citep{DBLP:journals/corr/abs-2311-08788}, Prometheus~\citep{DBLP:journals/corr/abs-2310-08491}, Auto-J~\citep{DBLP:journals/corr/abs-2310-05470}, InstructScore~\citep{DBLP:conf/emnlp/XuWPSFWL23} and TIGERScore~\citep{DBLP:journals/corr/abs-2310-00752}. Vers., Inde., Flex., Inte., and Open. represent versatility, independence, flexibility, interpretability, and open-source, respectively.}
\label{tab:1}
\end{table}

Overall, our main contributions are as follows:

\begin{itemize}
    \item We construct a large-scale NLG evaluation corpus, including about 0.5 million samples and 58 datasets across 9 NLG tasks, with detailed meta information, aspect criteria, and evaluations from both humans and GPT-4.
    \item We propose Themis, an LLM dedicated to NLG evaluation, which has been trained through our specific consistency and alignment methods and possesses versatility, independence, flexibility, and interpretability.
    \item Extensive experiments demonstrate the superior evaluation performance of Themis on common NLG tasks, as well as good generalization and robustness. Our model and relevant resource have been released in \href{https://github.com/PKU-ONELab/Themis}{https://github.com/PKU-ONELab/Themis} to facilitate related research.
    
\end{itemize}

\section{NLG-Eval Corpus}

Despite abundant data on NLG tasks, the corresponding high-quality evaluation data remains scarce and scattered due to the high cost of professional human annotations. Previous related methods either used a small amount of human annotations to train regression models, such as UniEval~\citep{DBLP:conf/emnlp/Zhong0YMJLZJH22} and X-Eval~\citep{DBLP:journals/corr/abs-2311-08788}, resulting in limited task coverage and a lack of evaluation analyses, or they entirely relied on LLMs to generate synthetic data, which raised reliability concerns~\citep{DBLP:journals/corr/abs-2402-12055}. To address this challenge, we formally defined the NLG evaluation task and clarified the involved elements. Subsequently, we surveyed a large number of existing related studies and compiled 58 evaluation datasets with human annotations across 9 NLG tasks, totaling 0.5 million samples. They have undergone our meticulous proofreading, and the missing but critical content has been supplemented, such as evaluation criteria. Additionally, we utilize the competent GPT-4 for supplementary evaluations, including analyses and ratings. The corpus has also been equipped with meta information, aiming to promote the development of related research.

\begin{figure}[t]
\centering
\includegraphics[width=0.482\textwidth]{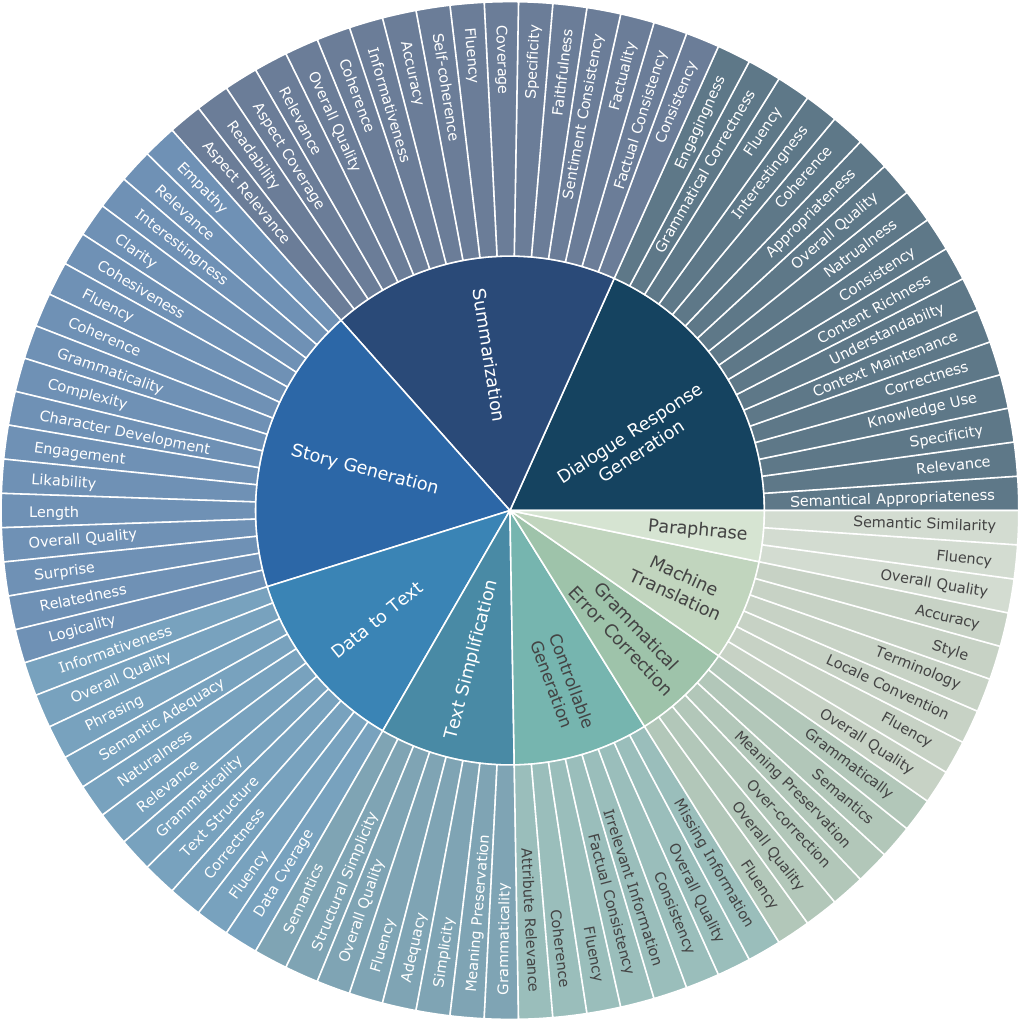}
\caption{The NLG evaluation tasks and corresponding evaluation aspects in our NLG-Eval corpus.}
\label{fig:2}
\end{figure}

\subsection{Definitions}

NLG tasks typically require language models to generate the corresponding output $t$ based on the given input $i$ and task requirements. The input in specific tasks may consist of multiple components; for example, in controlled generation, it is composed of the generation requirements and the constraint labels. NLG evaluation involves evaluating $t$ in conjunction with $i$ through the evaluation model or method $E$, using evaluation criteria $c$ and references $r$ if available. When they are absent, the evaluation defaults to general and reference-free evaluation, respectively. Evaluation results come in various forms, primarily focusing on rating $s$ and additional information $a$ such as analyses. The formalization of the evaluation process is as follows, with square brackets indicating optional elements:
\begin{equation*}
    s, [a] = E(t, i, [r], [c])
\end{equation*}
In the corpus construction, we collect and annotate the aforementioned elements for each sample.

\subsection{Data Collection}

To improve the coverage and diversity of the corpus, we conducted a comprehensive survey of related work on NLG tasks, including summarization, dialogue response generation, data-to-text, paraphrase, and so on. We collected datasets with human evaluation ratings and performed meticulous manual checking and filtering, ultimately obtaining 58 high-quality datasets across 9 different NLG tasks, which contain about 0.5 million samples. However, we discovered that many datasets, despite including evaluation aspects, lacked specific criteria or definitions, which could lead to confusion and ambiguity in evaluations. Such concerns have also been pointed out by~\citet{DBLP:conf/naacl/ZhouBTDSO22, DBLP:journals/corr/abs-2402-12055}. So we analyzed all these datasets, taking samples, human labels, and information from related papers into account, and manually supplementing the criteria. The different tasks included in our NLG-Eval corpus and their respective involved evaluation aspects are gathered and shown in Figure~\ref{fig:2}, with semantically similar aspects being merged. And more detailed statistics and information are presented in Appendix~\ref{sec:corpus}.

\subsection{GPT-4 Annotations}

Although human evaluation is often considered the most reliable, many datasets have sparse or inconsistent human annotations, leading to questions about their accuracy. Therefore, we utilized the superior language capabilities of GPT-4 to evaluate each sample in the corpus, as it has been shown to perform well on evaluation~\citep{DBLP:conf/nips/ZhengC00WZL0LXZ23}. GPT-4 can serve as an additional annotator to cross-verify with human evaluations, thereby improving reliability. On the other hand, it can provide evaluation analyses beyond just ratings and increase interpretability, which is lacking in human evaluation. Following conclusions and suggestions from~\citet{DBLP:conf/emnlp/ChiangL23, DBLP:journals/corr/abs-2312-16171}, we carefully designed instructions, ensuring that a concise and accurate analysis is provided before the rating. Each sample is evaluated with a temperature setting of 1, and 10 diverse evaluation results are obtained through multiple samplings, with more details included in Appendix~\ref{sec:prompt}. Additionally, our annotation and subsequent training did not use references because many datasets lack references or only have automatically constructed references that are unsatisfactory~\citep{DBLP:journals/corr/abs-2309-09558}. And the references are actually difficult to obtain in practice.

\section{Methodology}

Existing studies have highlighted that data quality is more crucial than quantity in LLM fine-tuning~\citep{DBLP:journals/corr/abs-2312-10302, DBLP:conf/nips/ZhouLX0SMMEYYZG23}. Therefore, we adopted empirical methods to improve the diversity and balance of data distribution and designed a multi-perspective consistency verification approach to obtain high-quality samples from the constructed corpus. Then they would be used for supervised fine-tuning based on open-source LLMs. To further enhance the evaluation capability and alignment of the model, we proposed a preference data construction and training method guided by evaluation ratings to improve the fine-tuned model.

\subsection{Diversity and Balance}

Similar to many studies on LLM training, we initially employed empirical approaches to sample data from the original corpus, ensuring adequate diversity and balanced distribution, which was generally considered beneficial for model training and generalization. To be specific, we treated the NLG task, evaluation aspect, and rating triplets as identifiers of data categories to cover as comprehensive a distribution as possible. And we sampled 100 pieces of data from each category while also considering the semantic diversity of the input texts involved. Given the insufficiency of data in certain categories and the requirement to introduce diversity in evaluation analyses and aspect criteria, we conducted multiple samplings of analyses differently for each category to further balance the data distribution, simultaneously rephrasing the evaluation aspect criteria. In the end, we obtained a dataset of approximately 67K for supervised fine-tuning. The detailed sampling method for both samples and evaluations will be introduced in the next subsection.

\subsection{Multi-perspective Consistency Verification}

During the sampling process, most categories actually include abundant samples, and each sample also involves multiple evaluations annotated by GPT-4. To take full advantage of such sufficient data resources and enhance the reliability of the training data, we verified the consistency of each sample and their evaluations from three different perspectives, ultimately obtaining the potentially high-quality data through screening.

\paragraph{Self-Consistency} When dealing with complex tasks like mathematical problems and logical reasoning, the chain of thought (CoT)~\citep{DBLP:conf/nips/Wei0SBIXCLZ22} method is always employed on LLMs to improve their performance. It involves generating intermediate steps before providing the final result, which generally leads to higher-quality responses. Furthermore, by performing multiple diverse samplings and selecting the most frequently occurring final result from the responses, the output can be more stable and accurate~\citep{DBLP:conf/iclr/0002WSLCNCZ23}. This process is referred to as self-consistency, which can also be regarded as a metric and calculated as the proportion of the most frequently occurring result to the total number of samplings. A higher self-consistency indicates greater certainty of the model itself and, thus, higher reliability of the responses to the corresponding samples.

In our evaluation annotation, GPT-4 was required to generate an analysis before assigning a rating, similar to a CoT process, which allows using self-consistency for data filtering. The formal definitions of the most frequently occurring rating $\hat{r}$ (called consistent rating) and \text{self-consistency} are as follows, where $n$ denotes the number of evaluations for a sample, and $r_i$ represents the rating of the $i$-th evaluation:
\begin{align*}
    \hat{r} = \arg \max_r & \sum_{i=1}^n \mathbbm{1}\left(r_i=r\right) \\
    \text{self-consistency} = & \frac{\sum_{i=1}^n \mathbbm{1}\left(r_i=\hat{r}\right)}{n}
\end{align*}
And we prioritized samples with high self-consistency and only retained their evaluations that include the consistent rating $\hat{r}$ to collect both potentially reliable samples and evaluations for the following process of verification.

\paragraph{Cross-Validation} The evaluation ratings serve as the most critical supervision signal in evaluation, making their accuracy paramount. Unlike previous related work that relied entirely on human labels or LLM-generated labels, we comprehensively considered the evaluations from both sources to conduct cross-validation. As for each sample, the corresponding consistent rating $\hat{r}$ is regarded as its rating from GPT-4 and scaled to the same range as the human rating. We prioritized samples where the two evaluation ratings were close, which reflected the high consistency between evaluations from humans and GPT-4 and also indicated the potential strong reliability of the samples.

\paragraph{Evaluation Inspection} \citet{DBLP:journals/corr/abs-2402-12055} indicated that although current LLMs like GPT-4 possess exciting evaluation capabilities, they may encounter issues of confusing aspect criteria, affecting the accuracy of their analyses and ratings. To address this, we propose two specific criteria for inspecting the evaluations themselves: consistency between the evaluation analysis and rating, and consistency between the evaluation analysis and aspect. We designed additional instructions to prompt GPT-4 to re-evaluate the candidate evaluations, whose details could be found in Appendix~\ref{sec:prompt}, and prioritized those being assessed as of good quality from both perspectives.

\subsection{Preference Alignment}

With the rapid rise of InstructGPT~\citep{DBLP:conf/nips/Ouyang0JAWMZASR22}, reinforcement learning from human feedback (RLHF), as one of the key technologies, has garnered wide attention and been applied in many subsequent LLMs with great success. So we have specifically modified DPO~\citep{DBLP:conf/nips/RafailovSMMEF23}, a commonly-used implementation of RLHF, based on our NLG evaluation scenarios, using evaluation ratings to guide the construction and training of preference data. The preference alignment was further conducted after the supervised fine-tuning of the model.

\subsubsection{Preference Construction}

Although the DPO method does not require training a separate reward model, it still necessitates preference pairs to convey preference information. Benefiting from multiple evaluations of each sample, we can directly regard the evaluation that matches the consistent rating $\hat{r}$ introduced in Section 3.2 as the chosen response and that does not as the rejected response to construct preference pairs. The quality gap between the chosen and rejected evaluations can be reflected by the difference in their evaluation ratings, as the closer the rating to $\hat{r}$, the more reliable the evaluation can be deemed. And our preference data is constructed based on the previously obtained training data without involving additional samples.

\subsubsection{Rating-guided DPO}

During vanilla DPO, we reparameterize the reward function $r$ using the policy as follows:
\[ r(x, y) = \beta \log \frac{\pi_\theta(y|x)}{\pi_\mathrm{ref}(y|x)} + \beta \log Z(x) \]
where $\pi_\theta$ is the policy model, $\pi_\mathrm{ref}$ is the reference model, typically a model undergone supervised fine-tuning, $\beta$ is the parameter controlling the degree of deviation between them, and $Z(x)$ is the partition function. Moreover, the Bradley-Terry (BT)~\citep{Bradley1952RankAO} model is employed to model preferences, yielding:
\begin{equation*}
    p(y_1 \succ y_2|x) = \sigma(r(x, y_1) - r(x, y_2)) 
\end{equation*}

For a pair of evaluation responses $(y_1, y_2)$, if we prefer $y_1$ to $y_2$, then the larger the evaluation rating difference between $y_1$ and $y_2$, the greater the preference difference $r(x, y_1) - r(x, y_2)$ should be. Therefore, to treat each preference pair more equally, we modify the BT model by subtracting a value proportional to the rating difference between $y_1$ and $y_2$ from $r(x, y_1) - r(x, y_2)$, thereby compensating for the prior preference difference caused by the rating difference, and then obtain:
\begin{align*}
    p^{*}(y_1 \succ y_2|x) =  \sigma( & r(x, y_1) - r(x, y_2) \\ 
    &- \alpha|R(y_1) - R(y_2)|)
\end{align*}
where $R(y)$ denotes the evaluation rating included in response $y$. Finally, the maximum likelihood of the new preference model serves as the optimization objective for our rating-guided DPO:
\begin{align*}
\mathcal{L}^{*}_{\mathrm{DPO}}(\pi_\theta;\pi_\mathrm{ref}) \hspace{30pt} & \\
 = -\mathbb{E}_{(x, y_c, y_r) \sim \mathcal{D}} \big[
\log \sigma \big( & \beta \log \frac{\pi_\theta(y_c|x)}{\pi_\mathrm{ref}(y_c|x)} \\ 
 - \beta \log \frac{\pi_\theta(y_r|x)}{\pi_\mathrm{ref}(y_r|x)} & - \alpha|R(y_c) - R(y_r)| \big)\big]
\end{align*}
where $(x, y_c, y_r)$ contains the input content, the chosen evaluation, and the rejected evaluation, respectively.

\section{Experiments}

\subsection{Benchmarks}

We conducted extensive experiments to evaluate our Themis across six commonly used evaluation datasets for different NLG tasks, including SummEval~\citep{DBLP:journals/tacl/FabbriKMXSR21} for summarization, Topical-Chat~\citep{DBLP:conf/interspeech/GopalakrishnanH19} for dialogue response generation, SFRES\&SFHOT~\citep{DBLP:conf/emnlp/WenGMSVY15} for data-to-text, QAGS~\citep{DBLP:conf/acl/WangCL20} for factuality, MANS~\citep{DBLP:conf/acl/GuanZFLDMFH20} for story generation, and WMT23~\citep{DBLP:conf/wmt/FreitagMLARTKBD23} for machine translation (zh-en). These datasets were collected into our NLG-Eval corpus but were excluded during the construction of our training data to ensure fairness.

\begin{table*}
\centering
\small
\setlength{\tabcolsep}{3pt}
\renewcommand{\arraystretch}{1.3}
\begin{tabular}{lccccccccccccc}
\toprule
\multirow{2}{*}{\raisebox{-3pt}{\textbf{Method}}} & \multicolumn{2}{c}{\textbf{SummEval}} & \multicolumn{2}{c}{\textbf{Topical-Chat}} & \multicolumn{2}{c}{\textbf{SFHOT\&RES}} & \multicolumn{2}{c}{\textbf{QAGS}} & \multicolumn{2}{c}{\textbf{MANS}} & \multicolumn{2}{c}{\textbf{WMT23}} & \multicolumn{1}{c}{\textbf{Average}} \\
\cmidrule(lr){2-3} \cmidrule(lr){4-5} \cmidrule(lr){6-7} \cmidrule(lr){8-9} \cmidrule(lr){10-11} \cmidrule(lr){12-13} \cmidrule(lr){14-14} 
 & $\rho$ & $\tau$ & $r$ & $\rho$ & $\rho$ & $\tau$ & $\rho$ & $\tau$ & $\rho$ & $\tau$ & $r$ & $\rho$ & $\rho$ \\
\midrule
\multicolumn{14}{l}{\emph{Traditional Metrics}} \\

BLEU$^{\dagger}$    & 0.075  & 0.057  & 0.356  & 0.388  & 0.024  &0.018& -  &  -    & 0.032  & 0.009  & -0.130  & 0.021  & -\\
ROUGE$^{\dagger}$    & 0.152  & 0.120  & 0.393  & 0.412  & 0.101  &0.076&   -    & -  & -0.002  & 0.156  & 0.081  & 0.151  & -\\
BARTScore    & 0.329  & 0.261  & 0.067  & 0.086  & 0.208  &0.156& 0.425  & 0.347  & 0.350  & 0.260  & 0.091  & 0.118  & 0.253\\ 
BERTScore$^{\dagger}$    & 0.231  & 0.182  & 0.388  & 0.394  & 0.139  &0.105& -  & -  & 0.285  & 0.163  & 0.123  & 0.219  & -\\
BLEURT$^{\dagger}$    & 0.152  & 0.118  & 0.384  & 0.388  & 0.244  &0.184&   -    &   -    & 0.138  & 0.221  & 0.163  & 0.263  & -\\
CometKiwi    & 0.228  & 0.180  & 0.353  & 0.340  & 0.251  &0.186& 0.094  & 0.074  & 0.251  & 0.176  & 0.413  & 0.343  & 0.251\\ 
UniEval$^{\dagger}$    & 0.474  & 0.377  & 0.533  & 0.577  & 0.282  &0.211&   -    &   -   & -  & -  & -  & -  & -\\

\midrule
\multicolumn{14}{l}{\emph{Prompting LLM}} \\
 G-Eval (GPT-3.5)    & 0.409  & 0.323  & 0.574  & 0.585  &   -     &-& 0.461  & 0.337  & -  & -  & -  & -  & - \\
G-Eval (GPT-4)    & \underline{0.523}  & 0.423  & 0.575  & 0.588  &   -    &-& 0.611  & 0.532  & -  & -  & -  & -  & -\\
GPT-3.5    & 0.416  & 0.340  & 0.592  & 0.578  & 0.306  &0.239& 0.431  & 0.356  & 0.328  & \underline{0.295}  & 0.388  & 0.347  & 0.401 \\ 
GPT-4    & 0.511  & \underline{0.423}  &  \textbf{0.770}  &  \textbf{0.746}  & \underline{0.320}  &\underline{0.260}& \underline{0.637}  & \underline{0.532}  & \underline{0.473}  & 0.260  &  \textbf{0.496}  &  \textbf{0.437}  & \underline{0.521}\\ 

\midrule
\multicolumn{14}{l}{\emph{Fine-tuned LLM}} \\
X-Eval$^{\dagger}$    & 0.480  & 0.362  & 0.539  & 0.605  & 0.303  &-& 0.578  &   -    & -  & -  & -  & -  & -\\
Prometheus-13B$^{\dagger}$    & 0.163  & 0.142  & 0.435  & 0.434  & 0.173  &0.142& -  & -  & 0.007  & 0.146  & 0.144  & 0.129  & -\\
Auto-J-13B    & 0.198  & 0.172  & 0.427  & 0.425  & 0.141  &0.120& 0.226  & 0.209  & 0.380  & 0.284  & 0.128  & 0.104  & 0.246\\
TIGERScore-13B    & 0.384  & 0.334  & 0.334  & 0.346  & 0.200  &0.175& 0.504  & 0.446  & 0.231  & 0.207  & 0.277  & 0.248  & 0.319 \\ 
InstructScore-7B$^{\dagger}$    & 0.258  & 0.226  & 0.269  & 0.241  & 0.247  &0.210&   -   & -  & 0.298  & 0.168  & 0.213  & 0.219  & - \\
\textbf{Themis-8B (ours)}    &  \textbf{0.553}  &  \textbf{0.499}  &  \underline{0.733}   & \underline{0.725}  &  \textbf{0.333}  &\textbf{0.284}&  \textbf{0.684}  &  \textbf{0.613}  &  \textbf{0.551}  &  \textbf{0.501}  &   \underline{0.431}    & \underline{0.405}  &  \textbf{0.542}\\ 

\bottomrule
\end{tabular}
\caption{The results of our Themis compared with different evaluation metrics and models on six different NLG tasks. ${\dagger}$ represents reference-based methods, while bold and underline indicate the first and second best results.}
\label{tab:2}
\end{table*}

\subsection{Experimental Settings}

We chose Llama-3-8B~\citep{llama3} for supervised fine-tuning and preference alignment in our main experiments, and more training details are described in Appendix~\ref{sec:training}. When assessing the evaluation capability of different models, we calculated the correlation between evaluation ratings from the model and humans using Pearson ($r$), Kendall ($\tau$), and Spearman ($\rho$) correlation coefficients. The specific calculations and the selection of correlation coefficients vary across different datasets, and we follow the common setups used in previous work. Furthermore, while some studies~\citep{DBLP:conf/emnlp/ChiangL23} employed a temperature setting of T = 1 to perform multiple sampling and aggregate the evaluation results, which may enhance performance, we believe it is costly and instable in practice and actually leads to the loss of evaluation analyses. Therefore, we tested our Themis with T = 0 and single sampling, while using their original settings for other methods. Given that most metrics and models do not support specific evaluation aspects, we calculate the results of different aspects using their overall scores, as in previous work.

\subsection{Baselines}

We experimented with existing representative and commonly-used evaluation metrics and models from three categories. Traditional metrics include BLEU, ROUGE~\citep{lin-2004-ROUGE}, BARTScore, BERTScore~\citep{DBLP:conf/iclr/ZhangKWWA20}, BLEURT~\citep{DBLP:conf/acl/SellamDP20}, CometKiwi~\citep{DBLP:conf/wmt/ReiTGZFMSGACLM22}, and UniEval. Methods of direct prompting LLMs include G-Eval and the evaluations with GPT-3.5 and GPT-4, implemented following~\citet{DBLP:conf/emnlp/ChiangL23}. Fine-tuned evaluation LLMs include X-Eval, Prometheus, Auto-J, TIGERScore, and InstructScore. Many of them are reference-based and therefore cannot be tested on the reference-free QAGS dataset.

\subsection{Main Results}

The main experimental results of our Themis compared with other baselines are shown in Table~\ref{tab:2}. We present the average correlation coefficients for each task, with the complete results demonstrated in Appendix~\ref{sec:6NLG}. Our Themis achieves the best overall evaluation performance and surpasses all larger LLMs. Furthermore, Themis also behaves better on each of six NLG tasks than other baselines, except GPT-4. It only outperforms Themis in dialogue response and translation evaluations, likely because GPT-4 has considerably stronger dialogue and multilingual capabilities than Llama-3-8B, which are almost impossible to improve during our limited training. And our Themis avoids the limitations of high costs and instability in proprietary LLMs and can be conveniently used offline. Moreover, although models such as BLEURT, X-Eval, and Prometheus conduct evaluation based on references, they still lag behind our reference-free Themis, showing its superiority.

\subsection{Ablation Study}

\begin{table}
\centering
\small
\setlength{\tabcolsep}{9pt}
\renewcommand{\arraystretch}{1.3}
\begin{tabular}{lccc}
\toprule

\textbf{Method} & \textbf{Avg $\rho$} & \textbf{Avg $\tau$} & \textbf{Avg $r$} \\
\midrule

BARTScore & 0.253 & 0.197 & 0.262 \\
CometKiwi & 0.251 & 0.195 & 0.272 \\
GPT-4  & 0.521 & 0.417 & 0.564 \\
TIGERScore-13B & 0.319 & 0.280 & 0.333 \\
Auto-J-13B & 0.246 & 0.211 & 0.262 \\

\textbf{Themis-8B (ours)} & \textbf{0.542} & \textbf{0.486} & \textbf{0.569} \\

\midrule 

\multicolumn{3}{l}{\emph{Data Sampling}} \\
\quad 100\% raw data  & 0.493 & 0.445 & 0.515\\
\quad 50\% raw data & 0.478 & 0.432 & 0.505\\ 
\quad 67K raw data & 0.476 & 0.428 & 0.507 \\

\midrule

\multicolumn{3}{l}{\emph{Preference Alignment}} \\

\quad vanilla DPO & 0.528 & 0.474 & 0.556\\
\quad without DPO & 0.508 & 0.456 & 0.534\\
\midrule

\multicolumn{3}{l}{\emph{Foundational Model}} \\

\quad Mistral-7B & 0.489 & 0.440 & 0.515 \\
\quad Llama-2-7B & 0.478 & 0.430 & 0.508 \\
\quad Llama-2-13B & 0.500 & 0.449 & 0.533 \\

\bottomrule
\end{tabular}
\caption{The results with different ablation settings.}
\label{tab:3}
\end{table}

We conducted ablation studies on our two main methods: consistency-based data sampling for supervised fine-tuning and rating-guided preference alignment. Additionally, considering the superior performance of Llama-3 among LLMs of the same size, we also experimented with other LLMs as the foundational model to verify the effectiveness of our approach. The average results for the six task datasets with different settings are shown in Table~\ref{tab:3}.

\noindent \textbf{Data Sampling} We experimented with approximately 67K raw data that has the same scale as the constructed data for supervised fine-tuning, and further trained the model using more randomly sampled data, namely half and all of the corpus data. The results show that although more data bring some improvements, the training cost increases significantly, and there is still a substantial performance drop compared to our current model.

\noindent \textbf{Preference Alignment} The model trained with our rating-guided DPO method exhibits progressive improvements over the model trained with vanilla DPO, and then the model that has only been fine-tuned. It indicates that preference alignment has certain effects on the training of NLG evaluation tasks, and our specific method is better suited for evaluation scenarios.

\noindent \textbf{Foundational Model} Although the evaluation models trained based on other LLMs do not perform as well as our current model, the models fine-tuned on Llama-2-7B still significantly outperformed TIGERScore and Auto-J, which use Llama-2-13B as the foundation model. These results demonstrate the effectiveness and necessity of the data sampling and alignment training methods we introduced.

\section{Reliability Analyses}

To analyze and investigate the reliability of our Themis, we consider two aspects: generalizability, which means the model can effectively evaluate unseen NLG tasks during training, and robustness, which means the model can accurately evaluate texts containing noises and perturbations.

\begin{table*}
\centering
\small
\setlength{\tabcolsep}{5pt}
\renewcommand{\arraystretch}{1.3}
\begin{tabular}{lccccccccccc}
\toprule
\multirow{2}{*}{\centering \textbf{Method}} & \multicolumn{5}{c}{\textbf{Instruction Following}} & \multicolumn{5}{c}{\textbf{Long form Question Answering}} & \multirow{2}{*}{\centering \textbf{Average $\rho$}} \\ 
\cmidrule(lr){2-6} \cmidrule(lr){7-11}
& \textbf{CLA} & \textbf{COM} & \textbf{COR} & \textbf{POL} & \textbf{REL} & \textbf{EU} & \textbf{FA} & \textbf{LC} & \textbf{OU} & \textbf{SL} & \\
\midrule

BARTScore & -0.053 & -0.040 & 0.063 & -0.038 & 0.025 & 0.098 & 0.174 & 0.360 & 0.296 & 0.464 & 0.135\\
CometKiwi & 0.345 & 0.464 & 0.349 & 0.223 & 0.473 & 0.155 & 0.128 & 0.210 & 0.138 & 0.134 & 0.262\\
GPT-3.5 & 0.257 & 0.488 & 0.433 & 0.239 & 0.401 & 0.612 & 0.318 & 0.443 & 0.212 & 0.562 & 0.396\\
GPT-4 & 0.257 & \textbf{0.653} & 0.573 & 0.338 & 0.455 & 0.767 & 0.431 & 0.237 & -0.004 & \textbf{0.711} & 0.442\\
Auto-J & \textbf{0.422} & 0.562 & 0.571 & \textbf{0.386} & \textbf{0.548} & 0.095 & 0.255 & 0.109 & 0.191 & 0.033 & 0.317\\
TIGERScore & 0.262 & 0.247 & 0.285 & 0.198 & 0.239 & -0.037 & 0.075 & 0.009 & 0.002 & -0.217 & 0.106\\
Themis (ours) & 0.414 & 0.381 & \textbf{0.685} & 0.349 & 0.500 & \textbf{0.835} & \textbf{0.538} & \textbf{0.700} & \textbf{0.365} & 0.684 & \textbf{0.545}\\

\bottomrule
\end{tabular}
\caption{The spearman correlation between evaluations from humans and different models on instruction-following and long-form question-answering tasks.}
\label{tab:4}
\end{table*}

\subsection{Unseen Tasks}

We tested our model on two common but unseen evaluation tasks during training: long-form question answering and instruction-following evaluation~\citep{DBLP:journals/corr/abs-2310-19740}. The former task involves five evaluation aspects: clarity (CLA), completeness (COM), correctness (COR), politeness (POL), and relevance (REL), while the latter involves example usage (EU), factual accuracy (FA), logical coherence (LC), overall usefulness (OU), and simple language (SL). As shown in Table~\ref{tab:4}, Themis outperforms both proprietary and open-source LLMs on the overall two tasks, demonstrating good generalization. And its performance is relatively well-balanced without particular weaknesses, unlike Auto-J, which performs well on instruction-following but poorly on long-form question-answering. Furthermore, some evaluation aspects are relatively open, which traditional reference-based metrics that do not support specific evaluation aspects cannot handle, highlighting the superiority of our Themis.

\subsection{Perturbation Tests}

\begin{table}
\centering
\small
\setlength{\tabcolsep}{3pt}
\renewcommand{\arraystretch}{1.3}
\begin{tabular}{lccccc}
\toprule

\textbf{Method} & \textbf{Dial} & \textbf{News} & \textbf{Para} & \textbf{Table} & \textbf{Average$\downarrow$} \\
\midrule

GPT-4 & 1.158 & 0.896 & 0.929 & 0.923 & 0.976 \\
Prometheus & 1.095 & 1.433 & 1.051 & 1.485 & 1.266 \\
Themis (ours) & 0.841 & 0.748 & 0.698 & 0.860 & 0.787 \\

\bottomrule
\end{tabular}
\caption{The results of perturbation tests, showing the changes in evaluation ratings from different models before and after perturbations.}
\label{tab:5}
\end{table}

\citet{DBLP:journals/corr/abs-2402-12055} has pointed out that both proprietary and open-source LLMs have certain issues with confusing evaluation aspects during NLG evaluation. Therefore, we conducted perturbation tests based on their methods to investigate the robustness of different evaluation models that support customized aspects. Specifically, we applied aspect-targeted perturbations on references in four tasks—news summarization, dialogue summarization, paraphrase generation, and data-to-text—focusing on fluency, coherence, informativeness, and consistency. The perturbations were designed to impact only the quality of the target aspect while leaving other aspects unaffected. We tested the models by comparing ratings on the perturbed texts with those on the original references, expecting little change and decreases in unaffected aspects. As shown in Table~\ref{tab:5}, Themis is closest to what is expected, with the smallest average decreases across the four tasks. And slight decreases may be caused by the fact that the designed pertubations do not entirely avoid affecting other aspects. On the other hand, the evaluation of perturbed texts may indeed be challenging, necessitating further exploration to enhance the model's capabilities.

\section{Related Works}

\subsection{Prompting LLMs for NLG Evaluation}

Previous research has evaluated text quality in various tasks by directly prompting proprietary LLMs (e.g. GPT-4) for scoring \cite{DBLP:conf/emnlp/LiuIXWXZ23,DBLP:conf/acl/ChiangL23,DBLP:conf/eamt/KocmiF23}, comparison \citep{DBLP:conf/eacl/LiusieMG24,DBLP:journals/corr/abs-2310-11593}, ranking \citep{DBLP:journals/corr/abs-2303-07610,DBLP:journals/corr/abs-2305-14239}, and error analysis \citep{DBLP:conf/wmt/KocmiF23,DBLP:journals/corr/abs-2303-13809}, surpassing traditional evaluation metrics in performance and providing significant flexibility.  \citet{DBLP:conf/eval4nlp/LeiterODGDE23} have experimented with various prompt formats and in-contextual example settings. Moreover, to better assess specific aspects, \citet{DBLP:conf/coling/LiuYHZHWDSZ24} have used LLMs to generate or improve the definitions of evaluation criteria for use in prompting, and  \citet{DBLP:journals/corr/abs-2312-10355} have had LLMs first evaluate other relevant aspects. Additionally, \citet{DBLP:conf/nlpcc/WuGSLJ23,DBLP:conf/nips/BaiY0LHWYZXLZLH23} have enhanced evaluative capabilities through interactions and role-playing among LLMs. Although these studies are compelling, they rely on proprietary models, which are costly and pose reproducibility issues.

\subsection{Fine-tuned LLM Evaluators}

In response to the issues with prompting LLMs, subsequent studies have shifted to fine-tuning relatively smaller open-source large models to create specialized evaluators \citep{DBLP:journals/corr/abs-2306-05087,DBLP:conf/emnlp/XuWPSFWL23,DBLP:journals/corr/abs-2310-08491,DBLP:journals/corr/abs-2308-04592,DBLP:journals/corr/abs-2310-00752,DBLP:journals/corr/abs-2310-05470,DBLP:journals/corr/abs-2311-18702,DBLP:journals/corr/abs-2310-17631,DBLP:journals/corr/abs-2311-08788}. Their approaches are generally similar: they use existing human evaluation data or synthetic data created by proprietary LLMs, such as GPT-4, as training data to fine-tune open-source LLMs like Llama \citep{DBLP:journals/corr/abs-2302-13971}. Their training modes vary in details, such as whether references are required and whether they support customized evaluation criteria. However, without specific measures, such fine-tuned evaluators seemed to contain issues similar to those in proprietary LLMs \citep{DBLP:journals/corr/abs-2402-12055}.

\section{Conclusions}

In this paper, we propose a large-scale and comprehensive NLG evaluation corpus and an LLM dedicated to NLG evaluation, Themis. The NLG-Eval corpus encompasses 9 common NLG tasks across 58 datasets, comprising about 0.5 million samples with evaluations from both human and GPT-4. Based on the constructed corpus, we propose a specialized multi-perspective consistency verification method to select high-quality supervised fine-tuning data, together with preference alignment guided by evaluation ratings, to train our Themis. It can be applied to various NLG tasks with flexible and interpretable evaluation in a reference-free manner. Extensive experiments demonstrate that our Themis performs well on different NLG tasks and can be generalized well to unseen tasks. Our model and resources have been released, and we hope they will promote further research in the field of NLG evaluation.

\section*{Limitations}

During the construction of our NLG-Eval corpus, we relied on annotations from GPT-4, resulting in considerable costs. However, it is worthwhile because our Themis, built upon this corpus and proposed training methods, has achieved great NLG evaluation capabilities. We have released the corresponding corpus and model to reduce the cost of future research and promote the development of related studies. Additionally, due to limited computational resources, we did not use larger and more powerful LLMs than Llama-3-8B as the foundational model for training, such as Llama-3-70B. Although larger models might offer stronger evaluation capabilities, their practical usage becomes inconvenient. Balancing model performance and usability requires further exploration in the future.

\section*{Acknowledgements}
This work was supported by Beijing Science and Technology Program (Z231100007423011), National Science Foundation of China (No. 62161160339), Ant Group Research Fund and Key Laboratory of Science, Technology and Standard in Press Industry (Key Laboratory of Intelligent Press Media Technology). We appreciate the anonymous reviewers for their helpful comments and all the people who have contributed to this work. Xiaojun Wan is the corresponding author.

\bibliography{custom}

\begin{thebibliography}{103}
\providecommand{\natexlab}[1]{#1}

\bibitem[{Alva{-}Manchego et~al.(2020)Alva{-}Manchego, Martin, Bordes, Scarton, Sagot, and Specia}]{DBLP:conf/acl/Alva-ManchegoMB20}
Fernando Alva{-}Manchego, Louis Martin, Antoine Bordes, Carolina Scarton, Beno{\^{\i}}t Sagot, and Lucia Specia. 2020.
\newblock \href {https://doi.org/10.18653/V1/2020.ACL-MAIN.424} {{ASSET:} {A} dataset for tuning and evaluation of sentence simplification models with multiple rewriting transformations}.
\newblock In \emph{Proceedings of the 58th Annual Meeting of the Association for Computational Linguistics, {ACL} 2020, Online, July 5-10, 2020}, pages 4668--4679. Association for Computational Linguistics.

\bibitem[{Alva{-}Manchego et~al.(2021)Alva{-}Manchego, Scarton, and Specia}]{DBLP:journals/coling/Alva-ManchegoSS21}
Fernando Alva{-}Manchego, Carolina Scarton, and Lucia Specia. 2021.
\newblock \href {https://doi.org/10.1162/COLI\_A\_00418} {The (un)suitability of automatic evaluation metrics for text simplification}.
\newblock \emph{Comput. Linguistics}, 47(4):861--889.

\bibitem[{Bai et~al.(2023)Bai, Ying, Cao, Lv, He, Wang, Yu, Zeng, Xiao, Lyu, Zhang, Li, and Hou}]{DBLP:conf/nips/BaiY0LHWYZXLZLH23}
Yushi Bai, Jiahao Ying, Yixin Cao, Xin Lv, Yuze He, Xiaozhi Wang, Jifan Yu, Kaisheng Zeng, Yijia Xiao, Haozhe Lyu, Jiayin Zhang, Juanzi Li, and Lei Hou. 2023.
\newblock \href {http://papers.nips.cc/paper\_files/paper/2023/hash/f64e55d03e2fe61aa4114e49cb654acb-Abstract-Datasets\_and\_Benchmarks.html} {Benchmarking foundation models with language-model-as-an-examiner}.
\newblock In \emph{Advances in Neural Information Processing Systems 36: Annual Conference on Neural Information Processing Systems 2023, NeurIPS 2023, New Orleans, LA, USA, December 10 - 16, 2023}.

\bibitem[{Bradley and Terry(1952)}]{Bradley1952RankAO}
Ralph~Allan Bradley and Milton~E. Terry. 1952.
\newblock \href {https://api.semanticscholar.org/CorpusID:125209808} {Rank analysis of incomplete block designs: I. the method of paired comparisons}.
\newblock \emph{Biometrika}, 39:324.

\bibitem[{Bsharat et~al.(2023)Bsharat, Myrzakhan, and Shen}]{DBLP:journals/corr/abs-2312-16171}
Sondos~Mahmoud Bsharat, Aidar Myrzakhan, and Zhiqiang Shen. 2023.
\newblock \href {https://doi.org/10.48550/ARXIV.2312.16171} {Principled instructions are all you need for questioning llama-1/2, {GPT-3.5/4}}.
\newblock \emph{CoRR}, abs/2312.16171.

\bibitem[{Castro~Ferreira et~al.(2020)Castro~Ferreira, Gardent, Ilinykh, van~der Lee, Mille, Moussallem, and Shimorina}]{castro-ferreira-etal-2020-2020}
Thiago Castro~Ferreira, Claire Gardent, Nikolai Ilinykh, Chris van~der Lee, Simon Mille, Diego Moussallem, and Anastasia Shimorina. 2020.
\newblock \href {https://aclanthology.org/2020.webnlg-1.7} {The 2020 bilingual, bi-directional {W}eb{NLG}+ shared task: Overview and evaluation results ({W}eb{NLG}+ 2020)}.
\newblock In \emph{Proceedings of the 3rd International Workshop on Natural Language Generation from the Semantic Web (WebNLG+)}, pages 55--76, Dublin, Ireland (Virtual). Association for Computational Linguistics.

\bibitem[{Chhun et~al.(2022)Chhun, Colombo, Suchanek, and Clavel}]{DBLP:conf/coling/ChhunCSC22}
Cyril Chhun, Pierre Colombo, Fabian~M. Suchanek, and Chlo{\'{e}} Clavel. 2022.
\newblock \href {https://aclanthology.org/2022.coling-1.509} {Of human criteria and automatic metrics: {A} benchmark of the evaluation of story generation}.
\newblock In \emph{Proceedings of the 29th International Conference on Computational Linguistics, {COLING} 2022, Gyeongju, Republic of Korea, October 12-17, 2022}, pages 5794--5836. International Committee on Computational Linguistics.

\bibitem[{Chiang and Lee(2023{\natexlab{a}})}]{DBLP:conf/acl/ChiangL23}
David~Cheng{-}Han Chiang and Hung{-}yi Lee. 2023{\natexlab{a}}.
\newblock \href {https://doi.org/10.18653/V1/2023.ACL-LONG.870} {Can large language models be an alternative to human evaluations?}
\newblock In \emph{Proceedings of the 61st Annual Meeting of the Association for Computational Linguistics (Volume 1: Long Papers), {ACL} 2023, Toronto, Canada, July 9-14, 2023}, pages 15607--15631. Association for Computational Linguistics.

\bibitem[{Chiang and Lee(2023{\natexlab{b}})}]{DBLP:conf/emnlp/ChiangL23}
David~Cheng{-}Han Chiang and Hung{-}yi Lee. 2023{\natexlab{b}}.
\newblock \href {https://doi.org/10.18653/V1/2023.FINDINGS-EMNLP.599} {A closer look into using large language models for automatic evaluation}.
\newblock In \emph{Findings of the Association for Computational Linguistics: {EMNLP} 2023, Singapore, December 6-10, 2023}, pages 8928--8942. Association for Computational Linguistics.

\bibitem[{Dathathri et~al.(2020)Dathathri, Madotto, Lan, Hung, Frank, Molino, Yosinski, and Liu}]{DBLP:conf/iclr/DathathriMLHFMY20}
Sumanth Dathathri, Andrea Madotto, Janice Lan, Jane Hung, Eric Frank, Piero Molino, Jason Yosinski, and Rosanne Liu. 2020.
\newblock \href {https://openreview.net/forum?id=H1edEyBKDS} {Plug and play language models: {A} simple approach to controlled text generation}.
\newblock In \emph{8th International Conference on Learning Representations, {ICLR} 2020, Addis Ababa, Ethiopia, April 26-30, 2020}. OpenReview.net.

\bibitem[{Dusek et~al.(2020)Dusek, Novikova, and Rieser}]{DBLP:journals/csl/DusekNR20}
Ondrej Dusek, Jekaterina Novikova, and Verena Rieser. 2020.
\newblock \href {https://doi.org/10.1016/J.CSL.2019.06.009} {Evaluating the state-of-the-art of end-to-end natural language generation: The {E2E} {NLG} challenge}.
\newblock \emph{Comput. Speech Lang.}, 59:123--156.

\bibitem[{Fabbri et~al.(2021)Fabbri, Kryscinski, McCann, Xiong, Socher, and Radev}]{DBLP:journals/tacl/FabbriKMXSR21}
Alexander~R. Fabbri, Wojciech Kryscinski, Bryan McCann, Caiming Xiong, Richard Socher, and Dragomir~R. Radev. 2021.
\newblock \href {https://doi.org/10.1162/TACL\_A\_00373} {Summeval: Re-evaluating summarization evaluation}.
\newblock \emph{Trans. Assoc. Comput. Linguistics}, 9:391--409.

\bibitem[{Freitag et~al.(2021)Freitag, Foster, Grangier, Ratnakar, Tan, and Macherey}]{DBLP:journals/tacl/FreitagFGRTM21}
Markus Freitag, George~F. Foster, David Grangier, Viresh Ratnakar, Qijun Tan, and Wolfgang Macherey. 2021.
\newblock \href {https://doi.org/10.1162/TACL\_A\_00437} {Experts, errors, and context: {A} large-scale study of human evaluation for machine translation}.
\newblock \emph{Trans. Assoc. Comput. Linguistics}, 9:1460--1474.

\bibitem[{Freitag et~al.(2023)Freitag, Mathur, Lo, Avramidis, Rei, Thompson, Kocmi, Blain, Deutsch, Stewart, Zerva, Castilho, Lavie, and Foster}]{DBLP:conf/wmt/FreitagMLARTKBD23}
Markus Freitag, Nitika Mathur, Chi{-}kiu Lo, Eleftherios Avramidis, Ricardo Rei, Brian Thompson, Tom Kocmi, Fr{\'{e}}d{\'{e}}ric Blain, Daniel Deutsch, Craig Stewart, Chrysoula Zerva, Sheila Castilho, Alon Lavie, and George~F. Foster. 2023.
\newblock \href {https://doi.org/10.18653/V1/2023.WMT-1.51} {Results of {WMT23} metrics shared task: Metrics might be guilty but references are not innocent}.
\newblock In \emph{Proceedings of the Eighth Conference on Machine Translation, {WMT} 2023, Singapore, December 6-7, 2023}, pages 578--628. Association for Computational Linguistics.

\bibitem[{Gao et~al.(2024)Gao, Hu, Ruan, Pu, and Wan}]{DBLP:journals/corr/abs-2402-01383}
Mingqi Gao, Xinyu Hu, Jie Ruan, Xiao Pu, and Xiaojun Wan. 2024.
\newblock \href {https://doi.org/10.48550/ARXIV.2402.01383} {Llm-based {NLG} evaluation: Current status and challenges}.
\newblock \emph{CoRR}, abs/2402.01383.

\bibitem[{Gao and Wan(2022)}]{DBLP:conf/naacl/Gao022}
Mingqi Gao and Xiaojun Wan. 2022.
\newblock \href {https://doi.org/10.18653/V1/2022.NAACL-MAIN.418} {Dialsummeval: Revisiting summarization evaluation for dialogues}.
\newblock In \emph{Proceedings of the 2022 Conference of the North American Chapter of the Association for Computational Linguistics: Human Language Technologies, {NAACL} 2022, Seattle, WA, United States, July 10-15, 2022}, pages 5693--5709. Association for Computational Linguistics.

\bibitem[{Gardent et~al.(2017)Gardent, Shimorina, Narayan, and Perez{-}Beltrachini}]{DBLP:conf/acl/GardentSNP17}
Claire Gardent, Anastasia Shimorina, Shashi Narayan, and Laura Perez{-}Beltrachini. 2017.
\newblock \href {https://doi.org/10.18653/V1/P17-1017} {Creating training corpora for {NLG} micro-planners}.
\newblock In \emph{Proceedings of the 55th Annual Meeting of the Association for Computational Linguistics, {ACL} 2017, Vancouver, Canada, July 30 - August 4, Volume 1: Long Papers}, pages 179--188. Association for Computational Linguistics.

\bibitem[{Gong and Mao(2023{\natexlab{a}})}]{DBLP:journals/corr/abs-2312-10355}
Peiyuan Gong and Jiaxin Mao. 2023{\natexlab{a}}.
\newblock \href {https://doi.org/10.48550/ARXIV.2312.10355} {Coascore: Chain-of-aspects prompting for {NLG} evaluation}.
\newblock \emph{CoRR}, abs/2312.10355.

\bibitem[{Gong and Mao(2023{\natexlab{b}})}]{coascore}
Peiyuan Gong and Jiaxin Mao. 2023{\natexlab{b}}.
\newblock \href {https://doi.org/10.48550/ARXIV.2312.10355} {Coascore: Chain-of-aspects prompting for {NLG} evaluation}.
\newblock \emph{CoRR}, abs/2312.10355.

\bibitem[{Gopalakrishnan et~al.(2019)Gopalakrishnan, Hedayatnia, Chen, Gottardi, Kwatra, Venkatesh, Gabriel, and Hakkani{-}T{\"{u}}r}]{DBLP:conf/interspeech/GopalakrishnanH19}
Karthik Gopalakrishnan, Behnam Hedayatnia, Qinlang Chen, Anna Gottardi, Sanjeev Kwatra, Anu Venkatesh, Raefer Gabriel, and Dilek Hakkani{-}T{\"{u}}r. 2019.
\newblock \href {https://doi.org/10.21437/INTERSPEECH.2019-3079} {Topical-chat: Towards knowledge-grounded open-domain conversations}.
\newblock In \emph{20th Annual Conference of the International Speech Communication Association, Interspeech 2019, Graz, Austria, September 15-19, 2019}, pages 1891--1895. {ISCA}.

\bibitem[{Grusky et~al.(2018)Grusky, Naaman, and Artzi}]{DBLP:conf/naacl/GruskyNA18}
Max Grusky, Mor Naaman, and Yoav Artzi. 2018.
\newblock \href {https://doi.org/10.18653/V1/N18-1065} {Newsroom: {A} dataset of 1.3 million summaries with diverse extractive strategies}.
\newblock In \emph{Proceedings of the 2018 Conference of the North American Chapter of the Association for Computational Linguistics: Human Language Technologies, {NAACL-HLT} 2018, New Orleans, Louisiana, USA, June 1-6, 2018, Volume 1 (Long Papers)}, pages 708--719. Association for Computational Linguistics.

\bibitem[{Guan et~al.(2021)Guan, Zhang, Feng, Liu, Ding, Mao, Fan, and Huang}]{DBLP:conf/acl/GuanZFLDMFH20}
Jian Guan, Zhexin Zhang, Zhuoer Feng, Zitao Liu, Wenbiao Ding, Xiaoxi Mao, Changjie Fan, and Minlie Huang. 2021.
\newblock \href {https://doi.org/10.18653/V1/2021.ACL-LONG.500} {Openmeva: {A} benchmark for evaluating open-ended story generation metrics}.
\newblock In \emph{Proceedings of the 59th Annual Meeting of the Association for Computational Linguistics and the 11th International Joint Conference on Natural Language Processing, {ACL/IJCNLP} 2021, (Volume 1: Long Papers), Virtual Event, August 1-6, 2021}, pages 6394--6407. Association for Computational Linguistics.

\bibitem[{Gupta et~al.(2019)Gupta, Mehri, Zhao, Pavel, Esk{\'{e}}nazi, and Bigham}]{DBLP:conf/sigdial/GuptaMZPEB19}
Prakhar Gupta, Shikib Mehri, Tiancheng Zhao, Amy Pavel, Maxine Esk{\'{e}}nazi, and Jeffrey~P. Bigham. 2019.
\newblock \href {https://doi.org/10.18653/V1/W19-5944} {Investigating evaluation of open-domain dialogue systems with human generated multiple references}.
\newblock In \emph{Proceedings of the 20th Annual SIGdial Meeting on Discourse and Dialogue, SIGdial 2019, Stockholm, Sweden, September 11-13, 2019}, pages 379--391. Association for Computational Linguistics.

\bibitem[{He et~al.(2023)He, Zhang, Wang, Kumar, Cho, Glass, and Tsvetkov}]{DBLP:conf/acl/HeZ0KCGT23}
Tianxing He, Jingyu Zhang, Tianle Wang, Sachin Kumar, Kyunghyun Cho, James~R. Glass, and Yulia Tsvetkov. 2023.
\newblock \href {https://doi.org/10.18653/V1/2023.ACL-LONG.674} {On the blind spots of model-based evaluation metrics for text generation}.
\newblock In \emph{Proceedings of the 61st Annual Meeting of the Association for Computational Linguistics (Volume 1: Long Papers), {ACL} 2023, Toronto, Canada, July 9-14, 2023}, pages 12067--12097. Association for Computational Linguistics.

\bibitem[{Hu et~al.(2019)Hu, Rudinger, Post, and Durme}]{DBLP:conf/aaai/HuRPD19}
J.~Edward Hu, Rachel Rudinger, Matt Post, and Benjamin~Van Durme. 2019.
\newblock \href {https://doi.org/10.1609/AAAI.V33I01.33016521} {{PARABANK:} monolingual bitext generation and sentential paraphrasing via lexically-constrained neural machine translation}.
\newblock In \emph{The Thirty-Third {AAAI} Conference on Artificial Intelligence, {AAAI} 2019, The Thirty-First Innovative Applications of Artificial Intelligence Conference, {IAAI} 2019, The Ninth {AAAI} Symposium on Educational Advances in Artificial Intelligence, {EAAI} 2019, Honolulu, Hawaii, USA, January 27 - February 1, 2019}, pages 6521--6528. {AAAI} Press.

\bibitem[{Hu et~al.(2024)Hu, Gao, Hu, Zhang, Chen, Xu, and Wan}]{DBLP:journals/corr/abs-2402-12055}
Xinyu Hu, Mingqi Gao, Sen Hu, Yang Zhang, Yicheng Chen, Teng Xu, and Xiaojun Wan. 2024.
\newblock \href {https://doi.org/10.48550/ARXIV.2402.12055} {Are llm-based evaluators confusing {NLG} quality criteria?}
\newblock \emph{CoRR}, abs/2402.12055.

\bibitem[{Huang et~al.(2020{\natexlab{a}})Huang, Cui, Yang, Bao, Wang, Xie, and Zhang}]{DBLP:conf/emnlp/HuangCYBWXZ20}
Dandan Huang, Leyang Cui, Sen Yang, Guangsheng Bao, Kun Wang, Jun Xie, and Yue Zhang. 2020{\natexlab{a}}.
\newblock \href {https://doi.org/10.18653/V1/2020.EMNLP-MAIN.33} {What have we achieved on text summarization?}
\newblock In \emph{Proceedings of the 2020 Conference on Empirical Methods in Natural Language Processing, {EMNLP} 2020, Online, November 16-20, 2020}, pages 446--469. Association for Computational Linguistics.

\bibitem[{Huang et~al.(2020{\natexlab{b}})Huang, Ye, Qin, Lin, and Liang}]{DBLP:conf/emnlp/HuangYQLL20}
Lishan Huang, Zheng Ye, Jinghui Qin, Liang Lin, and Xiaodan Liang. 2020{\natexlab{b}}.
\newblock \href {https://doi.org/10.18653/V1/2020.EMNLP-MAIN.742} {{GRADE:} automatic graph-enhanced coherence metric for evaluating open-domain dialogue systems}.
\newblock In \emph{Proceedings of the 2020 Conference on Empirical Methods in Natural Language Processing, {EMNLP} 2020, Online, November 16-20, 2020}, pages 9230--9240. Association for Computational Linguistics.

\bibitem[{Ji et~al.(2023)Ji, Gong, Peng, Ni, Sun, Pan, Ma, and Li}]{DBLP:journals/corr/abs-2303-07610}
Yunjie Ji, Yan Gong, Yiping Peng, Chao Ni, Peiyan Sun, Dongyu Pan, Baochang Ma, and Xiangang Li. 2023.
\newblock \href {https://doi.org/10.48550/ARXIV.2303.07610} {Exploring chatgpt's ability to rank content: {A} preliminary study on consistency with human preferences}.
\newblock \emph{CoRR}, abs/2303.07610.

\bibitem[{Jiang et~al.(2023)Jiang, Li, Zhang, Huang, Lin, and Chen}]{DBLP:journals/corr/abs-2310-00752}
Dongfu Jiang, Yishan Li, Ge~Zhang, Wenhao Huang, Bill~Yuchen Lin, and Wenhu Chen. 2023.
\newblock \href {https://doi.org/10.48550/ARXIV.2310.00752} {Tigerscore: Towards building explainable metric for all text generation tasks}.
\newblock \emph{CoRR}, abs/2310.00752.

\bibitem[{Ke et~al.(2023)Ke, Wen, Feng, Liu, Lei, Cheng, Wang, Zeng, Dong, Wang, Tang, and Huang}]{DBLP:journals/corr/abs-2311-18702}
Pei Ke, Bosi Wen, Zhuoer Feng, Xiao Liu, Xuanyu Lei, Jiale Cheng, Shengyuan Wang, Aohan Zeng, Yuxiao Dong, Hongning Wang, Jie Tang, and Minlie Huang. 2023.
\newblock \href {https://doi.org/10.48550/ARXIV.2311.18702} {Critiquellm: Scaling llm-as-critic for effective and explainable evaluation of large language model generation}.
\newblock \emph{CoRR}, abs/2311.18702.

\bibitem[{Ke et~al.(2022)Ke, Zhou, Lin, Li, Zhou, Zhu, and Huang}]{DBLP:conf/acl/KeZLLZZH22}
Pei Ke, Hao Zhou, Yankai Lin, Peng Li, Jie Zhou, Xiaoyan Zhu, and Minlie Huang. 2022.
\newblock \href {https://doi.org/10.18653/V1/2022.ACL-LONG.164} {Ctrleval: An unsupervised reference-free metric for evaluating controlled text generation}.
\newblock In \emph{Proceedings of the 60th Annual Meeting of the Association for Computational Linguistics (Volume 1: Long Papers), {ACL} 2022, Dublin, Ireland, May 22-27, 2022}, pages 2306--2319. Association for Computational Linguistics.

\bibitem[{Kim et~al.(2023)Kim, Shin, Cho, Jang, Longpre, Lee, Yun, Shin, Kim, Thorne, and Seo}]{DBLP:journals/corr/abs-2310-08491}
Seungone Kim, Jamin Shin, Yejin Cho, Joel Jang, Shayne Longpre, Hwaran Lee, Sangdoo Yun, Seongjin Shin, Sungdong Kim, James Thorne, and Minjoon Seo. 2023.
\newblock \href {https://doi.org/10.48550/ARXIV.2310.08491} {Prometheus: Inducing fine-grained evaluation capability in language models}.
\newblock \emph{CoRR}, abs/2310.08491.

\bibitem[{Kocmi and Federmann(2023{\natexlab{a}})}]{DBLP:conf/wmt/KocmiF23}
Tom Kocmi and Christian Federmann. 2023{\natexlab{a}}.
\newblock \href {https://doi.org/10.18653/V1/2023.WMT-1.64} {{GEMBA-MQM:} detecting translation quality error spans with {GPT-4}}.
\newblock In \emph{Proceedings of the Eighth Conference on Machine Translation, {WMT} 2023, Singapore, December 6-7, 2023}, pages 768--775. Association for Computational Linguistics.

\bibitem[{Kocmi and Federmann(2023{\natexlab{b}})}]{DBLP:conf/eamt/KocmiF23}
Tom Kocmi and Christian Federmann. 2023{\natexlab{b}}.
\newblock \href {https://aclanthology.org/2023.eamt-1.19} {Large language models are state-of-the-art evaluators of translation quality}.
\newblock In \emph{Proceedings of the 24th Annual Conference of the European Association for Machine Translation, {EAMT} 2023, Tampere, Finland, 12-15 June 2023}, pages 193--203. European Association for Machine Translation.

\bibitem[{Kong{-}Vega et~al.(2018)Kong{-}Vega, Shen, Wang, and D'Haro}]{DBLP:conf/iwsds/Kong-VegaSWD18}
Naomi Kong{-}Vega, Mingxin Shen, Mo~Wang, and Luis~Fernando D'Haro. 2018.
\newblock \href {https://doi.org/10.1007/978-981-13-9443-0\_32} {Subjective annotation and evaluation of three different chatbots {WOCHAT:} shared task report}.
\newblock In \emph{9th International Workshop on Spoken Dialogue System Technology, {IWSDS} 2018, Singapore, April 18-20, 2018}, volume 579 of \emph{Lecture Notes in Electrical Engineering}, pages 371--378. Springer.

\bibitem[{Kreutzer et~al.(2018)Kreutzer, Uyheng, and Riezler}]{DBLP:conf/acl/RiezlerKU18}
Julia Kreutzer, Joshua Uyheng, and Stefan Riezler. 2018.
\newblock \href {https://doi.org/10.18653/V1/P18-1165} {Reliability and learnability of human bandit feedback for sequence-to-sequence reinforcement learning}.
\newblock In \emph{Proceedings of the 56th Annual Meeting of the Association for Computational Linguistics, {ACL} 2018, Melbourne, Australia, July 15-20, 2018, Volume 1: Long Papers}, pages 1777--1788. Association for Computational Linguistics.

\bibitem[{Lee et~al.(2020)Lee, Lim, and Sedoc}]{DBLP:journals/corr/abs-2010-12741}
Seolhwa Lee, Heuiseok Lim, and Jo{\~{a}}o Sedoc. 2020.
\newblock \href {https://arxiv.org/abs/2010.12741} {An evaluation protocol for generative conversational systems}.
\newblock \emph{CoRR}, abs/2010.12741.

\bibitem[{Leiter et~al.(2023)Leiter, Opitz, Deutsch, Gao, Dror, and Eger}]{DBLP:conf/eval4nlp/LeiterODGDE23}
Christoph Leiter, Juri Opitz, Daniel Deutsch, Yang Gao, Rotem Dror, and Steffen Eger. 2023.
\newblock \href {https://aclanthology.org/2023.eval4nlp-1.10} {The eval4nlp 2023 shared task on prompting large language models as explainable metrics}.
\newblock In \emph{Proceedings of the 4th Workshop on Evaluation and Comparison of {NLP} Systems, Eval4NLP 2023, Bali, Indonesia, November 1, 2023}, pages 117--138. Association for Computational Linguistics.

\bibitem[{Li et~al.(2023{\natexlab{a}})Li, Sun, Yuan, Fan, Zhao, and Liu}]{DBLP:journals/corr/abs-2310-05470}
Junlong Li, Shichao Sun, Weizhe Yuan, Run{-}Ze Fan, Hai Zhao, and Pengfei Liu. 2023{\natexlab{a}}.
\newblock \href {https://doi.org/10.48550/ARXIV.2310.05470} {Generative judge for evaluating alignment}.
\newblock \emph{CoRR}, abs/2310.05470.

\bibitem[{Li et~al.(2023{\natexlab{b}})Li, Cui, Kong, and Bi}]{DBLP:journals/corr/abs-2310-19740}
Qintong Li, Leyang Cui, Lingpeng Kong, and Wei Bi. 2023{\natexlab{b}}.
\newblock \href {https://doi.org/10.48550/ARXIV.2310.19740} {Collaborative evaluation: Exploring the synergy of large language models and humans for open-ended generation evaluation}.
\newblock \emph{CoRR}, abs/2310.19740.

\bibitem[{Li et~al.(2023{\natexlab{c}})Li, Hui, Xia, Yang, Yang, Zhang, Si, Liu, Liu, Huang, and Li}]{DBLP:journals/corr/abs-2312-10302}
Yunshui Li, Binyuan Hui, Xiaobo Xia, Jiaxi Yang, Min Yang, Lei Zhang, Shuzheng Si, Junhao Liu, Tongliang Liu, Fei Huang, and Yongbin Li. 2023{\natexlab{c}}.
\newblock \href {https://doi.org/10.48550/ARXIV.2312.10302} {One shot learning as instruction data prospector for large language models}.
\newblock \emph{CoRR}, abs/2312.10302.

\bibitem[{Lin(2004)}]{lin-2004-ROUGE}
Chin-Yew Lin. 2004.
\newblock \href {https://aclanthology.org/W04-1013} {{ROUGE}: A package for automatic evaluation of summaries}.
\newblock In \emph{Text Summarization Branches Out}, pages 74--81, Barcelona, Spain. Association for Computational Linguistics.

\bibitem[{Liu et~al.(2023{\natexlab{a}})Liu, Shen, Xu, Cao, Cho, Kumar, Ghanadan, and Huang}]{DBLP:journals/corr/abs-2311-08788}
Minqian Liu, Ying Shen, Zhiyang Xu, Yixin Cao, Eunah Cho, Vaibhav Kumar, Reza Ghanadan, and Lifu Huang. 2023{\natexlab{a}}.
\newblock \href {https://doi.org/10.48550/ARXIV.2311.08788} {X-eval: Generalizable multi-aspect text evaluation via augmented instruction tuning with auxiliary evaluation aspects}.
\newblock \emph{CoRR}, abs/2311.08788.

\bibitem[{Liu et~al.(2023{\natexlab{b}})Liu, Iter, Xu, Wang, Xu, and Zhu}]{DBLP:conf/emnlp/LiuIXWXZ23}
Yang Liu, Dan Iter, Yichong Xu, Shuohang Wang, Ruochen Xu, and Chenguang Zhu. 2023{\natexlab{b}}.
\newblock \href {https://doi.org/10.18653/V1/2023.EMNLP-MAIN.153} {G-eval: {NLG} evaluation using gpt-4 with better human alignment}.
\newblock In \emph{Proceedings of the 2023 Conference on Empirical Methods in Natural Language Processing, {EMNLP} 2023, Singapore, December 6-10, 2023}, pages 2511--2522. Association for Computational Linguistics.

\bibitem[{Liu et~al.(2023{\natexlab{c}})Liu, Fabbri, Chen, Zhao, Han, Joty, Liu, Radev, Wu, and Cohan}]{DBLP:journals/corr/abs-2311-09184}
Yixin Liu, Alexander~R. Fabbri, Jiawen Chen, Yilun Zhao, Simeng Han, Shafiq Joty, Pengfei Liu, Dragomir Radev, Chien{-}Sheng Wu, and Arman Cohan. 2023{\natexlab{c}}.
\newblock \href {https://doi.org/10.48550/ARXIV.2311.09184} {Benchmarking generation and evaluation capabilities of large language models for instruction controllable summarization}.
\newblock \emph{CoRR}, abs/2311.09184.

\bibitem[{Liu et~al.(2023{\natexlab{d}})Liu, Fabbri, Liu, Radev, and Cohan}]{DBLP:journals/corr/abs-2305-14239}
Yixin Liu, Alexander~R. Fabbri, Pengfei Liu, Dragomir Radev, and Arman Cohan. 2023{\natexlab{d}}.
\newblock \href {https://doi.org/10.48550/ARXIV.2305.14239} {On learning to summarize with large language models as references}.
\newblock \emph{CoRR}, abs/2305.14239.

\bibitem[{Liu et~al.(2023{\natexlab{e}})Liu, Feng, Wang, Zhang, and Sch{\"u}tze}]{liu2023evaluate}
Yongkang Liu, Shi Feng, Daling Wang, Yifei Zhang, and Hinrich Sch{\"u}tze. 2023{\natexlab{e}}.
\newblock Evaluate what you can't evaluate: Unassessable generated responses quality.
\newblock \emph{arXiv preprint arXiv:2305.14658}.

\bibitem[{Liu et~al.(2024{\natexlab{a}})Liu, Yang, Huang, Zhang, Huang, Wei, Deng, Sun, and Zhang}]{DBLP:conf/coling/LiuYHZHWDSZ24}
Yuxuan Liu, Tianchi Yang, Shaohan Huang, Zihan Zhang, Haizhen Huang, Furu Wei, Weiwei Deng, Feng Sun, and Qi~Zhang. 2024{\natexlab{a}}.
\newblock \href {https://aclanthology.org/2024.lrec-main.237} {Calibrating llm-based evaluator}.
\newblock In \emph{Proceedings of the 2024 Joint International Conference on Computational Linguistics, Language Resources and Evaluation, {LREC/COLING} 2024, 20-25 May, 2024, Torino, Italy}, pages 2638--2656. {ELRA} and {ICCL}.

\bibitem[{Liu et~al.(2024{\natexlab{b}})Liu, Yang, Huang, Zhang, Huang, Wei, Deng, Sun, and Zhang}]{AUTOCALIBRATE}
Yuxuan Liu, Tianchi Yang, Shaohan Huang, Zihan Zhang, Haizhen Huang, Furu Wei, Weiwei Deng, Feng Sun, and Qi~Zhang. 2024{\natexlab{b}}.
\newblock \href {https://aclanthology.org/2024.lrec-main.237} {Calibrating llm-based evaluator}.
\newblock In \emph{Proceedings of the 2024 Joint International Conference on Computational Linguistics, Language Resources and Evaluation, {LREC/COLING} 2024, 20-25 May, 2024, Torino, Italy}, pages 2638--2656. {ELRA} and {ICCL}.

\bibitem[{Liu et~al.(2024{\natexlab{c}})Liu, Yang, Huang, Zhang, Huang, Wei, Deng, Sun, and Zhang}]{HD-eval}
Yuxuan Liu, Tianchi Yang, Shaohan Huang, Zihan Zhang, Haizhen Huang, Furu Wei, Weiwei Deng, Feng Sun, and Qi~Zhang. 2024{\natexlab{c}}.
\newblock \href {https://doi.org/10.48550/ARXIV.2402.15754} {Hd-eval: Aligning large language model evaluators through hierarchical criteria decomposition}.
\newblock \emph{CoRR}, abs/2402.15754.

\bibitem[{Liusie et~al.(2024)Liusie, Manakul, and Gales}]{DBLP:conf/eacl/LiusieMG24}
Adian Liusie, Potsawee Manakul, and Mark J.~F. Gales. 2024.
\newblock \href {https://aclanthology.org/2024.eacl-long.8} {{LLM} comparative assessment: Zero-shot {NLG} evaluation through pairwise comparisons using large language models}.
\newblock In \emph{Proceedings of the 18th Conference of the European Chapter of the Association for Computational Linguistics, {EACL} 2024 - Volume 1: Long Papers, St. Julian's, Malta, March 17-22, 2024}, pages 139--151. Association for Computational Linguistics.

\bibitem[{Lu et~al.(2023)Lu, Qiu, Ding, Xie, and Tao}]{DBLP:journals/corr/abs-2303-13809}
Qingyu Lu, Baopu Qiu, Liang Ding, Liping Xie, and Dacheng Tao. 2023.
\newblock \href {https://doi.org/10.48550/ARXIV.2303.13809} {Error analysis prompting enables human-like translation evaluation in large language models: {A} case study on chatgpt}.
\newblock \emph{CoRR}, abs/2303.13809.

\bibitem[{Maddela et~al.(2023)Maddela, Dou, Heineman, and Xu}]{DBLP:conf/acl/MaddelaDHX23}
Mounica Maddela, Yao Dou, David Heineman, and Wei Xu. 2023.
\newblock \href {https://doi.org/10.18653/V1/2023.ACL-LONG.905} {{LENS:} {A} learnable evaluation metric for text simplification}.
\newblock In \emph{Proceedings of the 61st Annual Meeting of the Association for Computational Linguistics (Volume 1: Long Papers), {ACL} 2023, Toronto, Canada, July 9-14, 2023}, pages 16383--16408. Association for Computational Linguistics.

\bibitem[{Mehri and Esk{\'{e}}nazi(2020)}]{DBLP:conf/sigdial/MehriE20}
Shikib Mehri and Maxine Esk{\'{e}}nazi. 2020.
\newblock \href {https://doi.org/10.18653/V1/2020.SIGDIAL-1.28} {Unsupervised evaluation of interactive dialog with dialogpt}.
\newblock In \emph{Proceedings of the 21th Annual Meeting of the Special Interest Group on Discourse and Dialogue, SIGdial 2020, 1st virtual meeting, July 1-3, 2020}, pages 225--235. Association for Computational Linguistics.

\bibitem[{Merdivan et~al.(2020)Merdivan, Singh, Hanke, Kropf, Holzinger, and Geist}]{merdivan2020human}
Erinc Merdivan, Deepika Singh, Sten Hanke, Johannes Kropf, Andreas Holzinger, and Matthieu Geist. 2020.
\newblock Human annotated dialogues dataset for natural conversational agents.
\newblock \emph{Applied Sciences}, 10(3):762.

\bibitem[{Meta(2024)}]{llama3}
Meta. 2024.
\newblock Introducing meta llama 3: The most capable openly available llm to date.
\newblock \url{https://ai.meta.com/blog/meta-llama-3/}.

\bibitem[{Napoles et~al.(2019)Napoles, Nadejde, and Tetreault}]{DBLP:journals/tacl/NapolesNT19}
Courtney Napoles, Maria Nadejde, and Joel~R. Tetreault. 2019.
\newblock \href {https://doi.org/10.1162/TACL\_A\_00282} {Enabling robust grammatical error correction in new domains: Datasets, metrics, and analyses}.
\newblock \emph{Trans. Assoc. Comput. Linguistics}, 7:551--566.

\bibitem[{Novikova et~al.(2018)Novikova, Dusek, and Rieser}]{DBLP:conf/naacl/NovikovaDR18}
Jekaterina Novikova, Ondrej Dusek, and Verena Rieser. 2018.
\newblock \href {https://doi.org/10.18653/V1/N18-2012} {Rankme: Reliable human ratings for natural language generation}.
\newblock In \emph{Proceedings of the 2018 Conference of the North American Chapter of the Association for Computational Linguistics: Human Language Technologies, NAACL-HLT, New Orleans, Louisiana, USA, June 1-6, 2018, Volume 2 (Short Papers)}, pages 72--78. Association for Computational Linguistics.

\bibitem[{Novikova et~al.(2016)Novikova, Lemon, and Rieser}]{DBLP:conf/inlg/NovikovaLR16}
Jekaterina Novikova, Oliver Lemon, and Verena Rieser. 2016.
\newblock \href {https://doi.org/10.18653/V1/W16-6644} {Crowd-sourcing {NLG} data: Pictures elicit better data}.
\newblock In \emph{{INLG} 2016 - Proceedings of the Ninth International Natural Language Generation Conference, September 5-8, 2016, Edinburgh, {UK}}, pages 265--273. The Association for Computer Linguistics.

\bibitem[{Ouyang et~al.(2022)Ouyang, Wu, Jiang, Almeida, Wainwright, Mishkin, Zhang, Agarwal, Slama, Ray, Schulman, Hilton, Kelton, Miller, Simens, Askell, Welinder, Christiano, Leike, and Lowe}]{DBLP:conf/nips/Ouyang0JAWMZASR22}
Long Ouyang, Jeffrey Wu, Xu~Jiang, Diogo Almeida, Carroll~L. Wainwright, Pamela Mishkin, Chong Zhang, Sandhini Agarwal, Katarina Slama, Alex Ray, John Schulman, Jacob Hilton, Fraser Kelton, Luke Miller, Maddie Simens, Amanda Askell, Peter Welinder, Paul~F. Christiano, Jan Leike, and Ryan Lowe. 2022.
\newblock \href {http://papers.nips.cc/paper\_files/paper/2022/hash/b1efde53be364a73914f58805a001731-Abstract-Conference.html} {Training language models to follow instructions with human feedback}.
\newblock In \emph{Advances in Neural Information Processing Systems 35: Annual Conference on Neural Information Processing Systems 2022, NeurIPS 2022, New Orleans, LA, USA, November 28 - December 9, 2022}.

\bibitem[{Pagnoni et~al.(2021)Pagnoni, Balachandran, and Tsvetkov}]{DBLP:conf/naacl/PagnoniBT21}
Artidoro Pagnoni, Vidhisha Balachandran, and Yulia Tsvetkov. 2021.
\newblock \href {https://doi.org/10.18653/V1/2021.NAACL-MAIN.383} {Understanding factuality in abstractive summarization with {FRANK:} {A} benchmark for factuality metrics}.
\newblock In \emph{Proceedings of the 2021 Conference of the North American Chapter of the Association for Computational Linguistics: Human Language Technologies, {NAACL-HLT} 2021, Online, June 6-11, 2021}, pages 4812--4829. Association for Computational Linguistics.

\bibitem[{Pang et~al.(2020)Pang, Nijkamp, Han, Zhou, Liu, and Tu}]{DBLP:conf/acl/PangNHZLT20}
Bo~Pang, Erik Nijkamp, Wenjuan Han, Linqi Zhou, Yixian Liu, and Kewei Tu. 2020.
\newblock \href {https://doi.org/10.18653/V1/2020.ACL-MAIN.333} {Towards holistic and automatic evaluation of open-domain dialogue generation}.
\newblock In \emph{Proceedings of the 58th Annual Meeting of the Association for Computational Linguistics, {ACL} 2020, Online, July 5-10, 2020}, pages 3619--3629. Association for Computational Linguistics.

\bibitem[{Papineni et~al.(2002)Papineni, Roukos, Ward, and Zhu}]{DBLP:conf/acl/PapineniRWZ02}
Kishore Papineni, Salim Roukos, Todd Ward, and Wei{-}Jing Zhu. 2002.
\newblock \href {https://doi.org/10.3115/1073083.1073135} {Bleu: a method for automatic evaluation of machine translation}.
\newblock In \emph{Proceedings of the 40th Annual Meeting of the Association for Computational Linguistics, July 6-12, 2002, Philadelphia, PA, {USA}}, pages 311--318. {ACL}.

\bibitem[{Pu et~al.(2023)Pu, Gao, and Wan}]{DBLP:journals/corr/abs-2309-09558}
Xiao Pu, Mingqi Gao, and Xiaojun Wan. 2023.
\newblock \href {https://doi.org/10.48550/ARXIV.2309.09558} {Summarization is (almost) dead}.
\newblock \emph{CoRR}, abs/2309.09558.

\bibitem[{Rafailov et~al.(2023)Rafailov, Sharma, Mitchell, Manning, Ermon, and Finn}]{DBLP:conf/nips/RafailovSMMEF23}
Rafael Rafailov, Archit Sharma, Eric Mitchell, Christopher~D. Manning, Stefano Ermon, and Chelsea Finn. 2023.
\newblock \href {http://papers.nips.cc/paper\_files/paper/2023/hash/a85b405ed65c6477a4fe8302b5e06ce7-Abstract-Conference.html} {Direct preference optimization: Your language model is secretly a reward model}.
\newblock In \emph{Advances in Neural Information Processing Systems 36: Annual Conference on Neural Information Processing Systems 2023, NeurIPS 2023, New Orleans, LA, USA, December 10 - 16, 2023}.

\bibitem[{Rei et~al.(2020)Rei, Stewart, Farinha, and Lavie}]{DBLP:conf/emnlp/ReiSFL20}
Ricardo Rei, Craig Stewart, Ana~C. Farinha, and Alon Lavie. 2020.
\newblock \href {https://doi.org/10.18653/V1/2020.EMNLP-MAIN.213} {{COMET:} {A} neural framework for {MT} evaluation}.
\newblock In \emph{Proceedings of the 2020 Conference on Empirical Methods in Natural Language Processing, {EMNLP} 2020, Online, November 16-20, 2020}, pages 2685--2702. Association for Computational Linguistics.

\bibitem[{Rei et~al.(2022)Rei, Treviso, Guerreiro, Zerva, Farinha, Maroti, de~Souza, Glushkova, Alves, Coheur, Lavie, and Martins}]{DBLP:conf/wmt/ReiTGZFMSGACLM22}
Ricardo Rei, Marcos~V. Treviso, Nuno~Miguel Guerreiro, Chrysoula Zerva, Ana~C. Farinha, Christine Maroti, Jos{\'{e}} G.~C. de~Souza, Taisiya Glushkova, Duarte~M. Alves, Lu{\'{\i}}sa Coheur, Alon Lavie, and Andr{\'{e}} F.~T. Martins. 2022.
\newblock \href {https://aclanthology.org/2022.wmt-1.60} {Cometkiwi: Ist-unbabel 2022 submission for the quality estimation shared task}.
\newblock In \emph{Proceedings of the Seventh Conference on Machine Translation, {WMT} 2022, Abu Dhabi, United Arab Emirates (Hybrid), December 7-8, 2022}, pages 634--645. Association for Computational Linguistics.

\bibitem[{Schwarzer et~al.(2021)Schwarzer, Tanprasert, and Kauchak}]{schwarzer-etal-2021-improving}
Max Schwarzer, Teerapaun Tanprasert, and David Kauchak. 2021.
\newblock \href {https://doi.org/10.18653/v1/2021.textgraphs-1.10} {Improving human text simplification with sentence fusion}.
\newblock In \emph{Proceedings of the Fifteenth Workshop on Graph-Based Methods for Natural Language Processing (TextGraphs-15)}, pages 106--114, Mexico City, Mexico. Association for Computational Linguistics.

\bibitem[{Scialom et~al.(2021)Scialom, Martin, Staiano, de~la Clergerie, and Sagot}]{DBLP:journals/corr/abs-2104-07560}
Thomas Scialom, Louis Martin, Jacopo Staiano, {\'{E}}ric~Villemonte de~la Clergerie, and Beno{\^{\i}}t Sagot. 2021.
\newblock \href {https://arxiv.org/abs/2104.07560} {Rethinking automatic evaluation in sentence simplification}.
\newblock \emph{CoRR}, abs/2104.07560.

\bibitem[{Sedoc et~al.(2019)Sedoc, Ippolito, Kirubarajan, Thirani, Ungar, and Callison{-}Burch}]{DBLP:conf/naacl/SedocIKTUC19}
Jo{\~{a}}o Sedoc, Daphne Ippolito, Arun Kirubarajan, Jai Thirani, Lyle~H. Ungar, and Chris Callison{-}Burch. 2019.
\newblock \href {https://doi.org/10.18653/V1/N19-4011} {Chateval: {A} tool for chatbot evaluation}.
\newblock In \emph{Proceedings of the 2019 Conference of the North American Chapter of the Association for Computational Linguistics: Human Language Technologies, {NAACL-HLT} 2019, Minneapolis, MN, USA, June 2-7, 2019, Demonstrations}, pages 60--65. Association for Computational Linguistics.

\bibitem[{Sellam et~al.(2020)Sellam, Das, and Parikh}]{DBLP:conf/acl/SellamDP20}
Thibault Sellam, Dipanjan Das, and Ankur~P. Parikh. 2020.
\newblock \href {https://doi.org/10.18653/V1/2020.ACL-MAIN.704} {{BLEURT:} learning robust metrics for text generation}.
\newblock In \emph{Proceedings of the 58th Annual Meeting of the Association for Computational Linguistics, {ACL} 2020, Online, July 5-10, 2020}, pages 7881--7892. Association for Computational Linguistics.

\bibitem[{Shen et~al.(2022)Shen, Liu, Jiang, and Shi}]{DBLP:conf/emnlp/ShenLJ022}
Lingfeng Shen, Lemao Liu, Haiyun Jiang, and Shuming Shi. 2022.
\newblock \href {https://doi.org/10.18653/V1/2022.EMNLP-MAIN.208} {On the evaluation metrics for paraphrase generation}.
\newblock In \emph{Proceedings of the 2022 Conference on Empirical Methods in Natural Language Processing, {EMNLP} 2022, Abu Dhabi, United Arab Emirates, December 7-11, 2022}, pages 3178--3190. Association for Computational Linguistics.

\bibitem[{Shen and Wan(2023)}]{DBLP:journals/corr/abs-2310-18122}
Yuchen Shen and Xiaojun Wan. 2023.
\newblock \href {https://doi.org/10.48550/ARXIV.2310.18122} {Opinsummeval: Revisiting automated evaluation for opinion summarization}.
\newblock \emph{CoRR}, abs/2310.18122.

\bibitem[{Siledar et~al.(2024)Siledar, Nath, Muddu, Rangaraju, Nath, Bhattacharyya, Banerjee, Patil, Singh, Chelliah, and Garera}]{DBLP:journals/corr/abs-2402-11683}
Tejpalsingh Siledar, Swaroop Nath, Sankara Sri Raghava~Ravindra Muddu, Rupasai Rangaraju, Swaprava Nath, Pushpak Bhattacharyya, Suman Banerjee, Amey Patil, Sudhanshu~Shekhar Singh, Muthusamy Chelliah, and Nikesh Garera. 2024.
\newblock \href {https://doi.org/10.48550/ARXIV.2402.11683} {One prompt to rule them all: Llms for opinion summary evaluation}.
\newblock \emph{CoRR}, abs/2402.11683.

\bibitem[{Sottana et~al.(2023)Sottana, Liang, Zou, and Yuan}]{DBLP:conf/emnlp/SottanaLZY23}
Andrea Sottana, Bin Liang, Kai Zou, and Zheng Yuan. 2023.
\newblock \href {https://doi.org/10.18653/V1/2023.EMNLP-MAIN.543} {Evaluation metrics in the era of {GPT-4:} reliably evaluating large language models on sequence to sequence tasks}.
\newblock In \emph{Proceedings of the 2023 Conference on Empirical Methods in Natural Language Processing, {EMNLP} 2023, Singapore, December 6-10, 2023}, pages 8776--8788. Association for Computational Linguistics.

\bibitem[{Stiennon et~al.(2020)Stiennon, Ouyang, Wu, Ziegler, Lowe, Voss, Radford, Amodei, and Christiano}]{DBLP:journals/corr/abs-2009-01325}
Nisan Stiennon, Long Ouyang, Jeff Wu, Daniel~M. Ziegler, Ryan Lowe, Chelsea Voss, Alec Radford, Dario Amodei, and Paul~F. Christiano. 2020.
\newblock \href {https://arxiv.org/abs/2009.01325} {Learning to summarize from human feedback}.
\newblock \emph{CoRR}, abs/2009.01325.

\bibitem[{Sulem et~al.(2018{\natexlab{a}})Sulem, Abend, and Rappoport}]{DBLP:conf/emnlp/SulemAR18}
Elior Sulem, Omri Abend, and Ari Rappoport. 2018{\natexlab{a}}.
\newblock \href {https://doi.org/10.18653/V1/D18-1081} {{BLEU} is not suitable for the evaluation of text simplification}.
\newblock In \emph{Proceedings of the 2018 Conference on Empirical Methods in Natural Language Processing, Brussels, Belgium, October 31 - November 4, 2018}, pages 738--744. Association for Computational Linguistics.

\bibitem[{Sulem et~al.(2018{\natexlab{b}})Sulem, Abend, and Rappoport}]{DBLP:conf/naacl/SulemAR18}
Elior Sulem, Omri Abend, and Ari Rappoport. 2018{\natexlab{b}}.
\newblock \href {https://doi.org/10.18653/V1/N18-1063} {Semantic structural evaluation for text simplification}.
\newblock In \emph{Proceedings of the 2018 Conference of the North American Chapter of the Association for Computational Linguistics: Human Language Technologies, {NAACL-HLT} 2018, New Orleans, Louisiana, USA, June 1-6, 2018, Volume 1 (Long Papers)}, pages 685--696. Association for Computational Linguistics.

\bibitem[{Sulem et~al.(2018{\natexlab{c}})Sulem, Abend, and Rappoport}]{DBLP:conf/acl/AbendRS18}
Elior Sulem, Omri Abend, and Ari Rappoport. 2018{\natexlab{c}}.
\newblock \href {https://doi.org/10.18653/V1/P18-1016} {Simple and effective text simplification using semantic and neural methods}.
\newblock In \emph{Proceedings of the 56th Annual Meeting of the Association for Computational Linguistics, {ACL} 2018, Melbourne, Australia, July 15-20, 2018, Volume 1: Long Papers}, pages 162--173. Association for Computational Linguistics.

\bibitem[{Touvron et~al.(2023)Touvron, Lavril, Izacard, Martinet, Lachaux, Lacroix, Rozi{\`{e}}re, Goyal, Hambro, Azhar, Rodriguez, Joulin, Grave, and Lample}]{DBLP:journals/corr/abs-2302-13971}
Hugo Touvron, Thibaut Lavril, Gautier Izacard, Xavier Martinet, Marie{-}Anne Lachaux, Timoth{\'{e}}e Lacroix, Baptiste Rozi{\`{e}}re, Naman Goyal, Eric Hambro, Faisal Azhar, Aur{\'{e}}lien Rodriguez, Armand Joulin, Edouard Grave, and Guillaume Lample. 2023.
\newblock \href {https://doi.org/10.48550/ARXIV.2302.13971} {Llama: Open and efficient foundation language models}.
\newblock \emph{CoRR}, abs/2302.13971.

\bibitem[{Wang et~al.(2020)Wang, Cho, and Lewis}]{DBLP:conf/acl/WangCL20}
Alex Wang, Kyunghyun Cho, and Mike Lewis. 2020.
\newblock \href {https://doi.org/10.18653/V1/2020.ACL-MAIN.450} {Asking and answering questions to evaluate the factual consistency of summaries}.
\newblock In \emph{Proceedings of the 58th Annual Meeting of the Association for Computational Linguistics, {ACL} 2020, Online, July 5-10, 2020}, pages 5008--5020. Association for Computational Linguistics.

\bibitem[{Wang et~al.(2023{\natexlab{a}})Wang, Liang, Meng, Shi, Li, Xu, Qu, and Zhou}]{DBLP:journals/corr/abs-2303-04048}
Jiaan Wang, Yunlong Liang, Fandong Meng, Haoxiang Shi, Zhixu Li, Jinan Xu, Jianfeng Qu, and Jie Zhou. 2023{\natexlab{a}}.
\newblock \href {https://doi.org/10.48550/ARXIV.2303.04048} {Is chatgpt a good {NLG} evaluator? {A} preliminary study}.
\newblock \emph{CoRR}, abs/2303.04048.

\bibitem[{Wang et~al.(2023{\natexlab{b}})Wang, Yu, Tan, O'Brien, Pasunuru, Dwivedi{-}Yu, Golovneva, Zettlemoyer, Fazel{-}Zarandi, and Celikyilmaz}]{DBLP:journals/corr/abs-2308-04592}
Tianlu Wang, Ping Yu, Xiaoqing~Ellen Tan, Sean O'Brien, Ramakanth Pasunuru, Jane Dwivedi{-}Yu, Olga Golovneva, Luke Zettlemoyer, Maryam Fazel{-}Zarandi, and Asli Celikyilmaz. 2023{\natexlab{b}}.
\newblock \href {https://doi.org/10.48550/ARXIV.2308.04592} {Shepherd: {A} critic for language model generation}.
\newblock \emph{CoRR}, abs/2308.04592.

\bibitem[{Wang et~al.(2023{\natexlab{c}})Wang, Wei, Schuurmans, Le, Chi, Narang, Chowdhery, and Zhou}]{DBLP:conf/iclr/0002WSLCNCZ23}
Xuezhi Wang, Jason Wei, Dale Schuurmans, Quoc~V. Le, Ed~H. Chi, Sharan Narang, Aakanksha Chowdhery, and Denny Zhou. 2023{\natexlab{c}}.
\newblock \href {https://openreview.net/pdf?id=1PL1NIMMrw} {Self-consistency improves chain of thought reasoning in language models}.
\newblock In \emph{The Eleventh International Conference on Learning Representations, {ICLR} 2023, Kigali, Rwanda, May 1-5, 2023}. OpenReview.net.

\bibitem[{Wang et~al.(2023{\natexlab{d}})Wang, Jiang, Zhang, Li, Liang, Mei, and Bendersky}]{DBLP:journals/corr/abs-2310-11593}
Yaqing Wang, Jiepu Jiang, Mingyang Zhang, Cheng Li, Yi~Liang, Qiaozhu Mei, and Michael Bendersky. 2023{\natexlab{d}}.
\newblock \href {https://doi.org/10.48550/ARXIV.2310.11593} {Automated evaluation of personalized text generation using large language models}.
\newblock \emph{CoRR}, abs/2310.11593.

\bibitem[{Wang et~al.(2023{\natexlab{e}})Wang, Yu, Zeng, Yang, Wang, Chen, Jiang, Xie, Wang, Xie, Ye, Zhang, and Zhang}]{DBLP:journals/corr/abs-2306-05087}
Yidong Wang, Zhuohao Yu, Zhengran Zeng, Linyi Yang, Cunxiang Wang, Hao Chen, Chaoya Jiang, Rui Xie, Jindong Wang, Xing Xie, Wei Ye, Shikun Zhang, and Yue Zhang. 2023{\natexlab{e}}.
\newblock \href {https://doi.org/10.48550/ARXIV.2306.05087} {Pandalm: An automatic evaluation benchmark for {LLM} instruction tuning optimization}.
\newblock \emph{CoRR}, abs/2306.05087.

\bibitem[{Wei et~al.(2022)Wei, Wang, Schuurmans, Bosma, Ichter, Xia, Chi, Le, and Zhou}]{DBLP:conf/nips/Wei0SBIXCLZ22}
Jason Wei, Xuezhi Wang, Dale Schuurmans, Maarten Bosma, Brian Ichter, Fei Xia, Ed~H. Chi, Quoc~V. Le, and Denny Zhou. 2022.
\newblock \href {http://papers.nips.cc/paper\_files/paper/2022/hash/9d5609613524ecf4f15af0f7b31abca4-Abstract-Conference.html} {Chain-of-thought prompting elicits reasoning in large language models}.
\newblock In \emph{Advances in Neural Information Processing Systems 35: Annual Conference on Neural Information Processing Systems 2022, NeurIPS 2022, New Orleans, LA, USA, November 28 - December 9, 2022}.

\bibitem[{Wen et~al.(2015)Wen, Gasic, Mrksic, Su, Vandyke, and Young}]{DBLP:conf/emnlp/WenGMSVY15}
Tsung{-}Hsien Wen, Milica Gasic, Nikola Mrksic, Pei{-}hao Su, David Vandyke, and Steve~J. Young. 2015.
\newblock \href {https://doi.org/10.18653/V1/D15-1199} {Semantically conditioned lstm-based natural language generation for spoken dialogue systems}.
\newblock In \emph{Proceedings of the 2015 Conference on Empirical Methods in Natural Language Processing, {EMNLP} 2015, Lisbon, Portugal, September 17-21, 2015}, pages 1711--1721. The Association for Computational Linguistics.

\bibitem[{Wu et~al.(2023)Wu, Gong, Shou, Liang, and Jiang}]{DBLP:conf/nlpcc/WuGSLJ23}
Ning Wu, Ming Gong, Linjun Shou, Shining Liang, and Daxin Jiang. 2023.
\newblock \href {https://doi.org/10.1007/978-3-031-44693-1\_54} {Large language models are diverse role-players for summarization evaluation}.
\newblock In \emph{Natural Language Processing and Chinese Computing - 12th National {CCF} Conference, {NLPCC} 2023, Foshan, China, October 12-15, 2023, Proceedings, Part {I}}, volume 14302 of \emph{Lecture Notes in Computer Science}, pages 695--707. Springer.

\bibitem[{Xie et~al.(2023)Xie, Cohn, and Lau}]{DBLP:conf/inlg/XieCL23}
Zhuohan Xie, Trevor Cohn, and Jey~Han Lau. 2023.
\newblock \href {https://doi.org/10.18653/V1/2023.INLG-MAIN.23} {The next chapter: {A} study of large language models in storytelling}.
\newblock In \emph{Proceedings of the 16th International Natural Language Generation Conference, {INLG} 2023, Prague, Czechia, September 11 - 15, 2023}, pages 323--351. Association for Computational Linguistics.

\bibitem[{Xu et~al.(2023)Xu, Wang, Pan, Song, Freitag, Wang, and Li}]{DBLP:conf/emnlp/XuWPSFWL23}
Wenda Xu, Danqing Wang, Liangming Pan, Zhenqiao Song, Markus Freitag, William Wang, and Lei Li. 2023.
\newblock \href {https://doi.org/10.18653/V1/2023.EMNLP-MAIN.365} {{INSTRUCTSCORE:} towards explainable text generation evaluation with automatic feedback}.
\newblock In \emph{Proceedings of the 2023 Conference on Empirical Methods in Natural Language Processing, {EMNLP} 2023, Singapore, December 6-10, 2023}, pages 5967--5994. Association for Computational Linguistics.

\bibitem[{Yang and Klein(2021)}]{DBLP:conf/naacl/YangK21}
Kevin Yang and Dan Klein. 2021.
\newblock \href {https://doi.org/10.18653/V1/2021.NAACL-MAIN.276} {{FUDGE:} controlled text generation with future discriminators}.
\newblock In \emph{Proceedings of the 2021 Conference of the North American Chapter of the Association for Computational Linguistics: Human Language Technologies, {NAACL-HLT} 2021, Online, June 6-11, 2021}, pages 3511--3535. Association for Computational Linguistics.

\bibitem[{Yoshimura et~al.(2020)Yoshimura, Kaneko, Kajiwara, and Komachi}]{DBLP:conf/coling/YoshimuraKKK20}
Ryoma Yoshimura, Masahiro Kaneko, Tomoyuki Kajiwara, and Mamoru Komachi. 2020.
\newblock \href {https://doi.org/10.18653/V1/2020.COLING-MAIN.573} {{SOME:} reference-less sub-metrics optimized for manual evaluations of grammatical error correction}.
\newblock In \emph{Proceedings of the 28th International Conference on Computational Linguistics, {COLING} 2020, Barcelona, Spain (Online), December 8-13, 2020}, pages 6516--6522. International Committee on Computational Linguistics.

\bibitem[{Yuan et~al.(2021)Yuan, Neubig, and Liu}]{DBLP:conf/nips/YuanNL21}
Weizhe Yuan, Graham Neubig, and Pengfei Liu. 2021.
\newblock \href {https://proceedings.neurips.cc/paper/2021/hash/e4d2b6e6fdeca3e60e0f1a62fee3d9dd-Abstract.html} {Bartscore: Evaluating generated text as text generation}.
\newblock In \emph{Advances in Neural Information Processing Systems 34: Annual Conference on Neural Information Processing Systems 2021, NeurIPS 2021, December 6-14, 2021, virtual}, pages 27263--27277.

\bibitem[{Zhang et~al.(2021)Zhang, Sedoc, D'Haro, Banchs, and Rudnicky}]{DBLP:journals/corr/abs-2111-02110}
Chen Zhang, Jo{\~{a}}o Sedoc, Luis~Fernando D'Haro, Rafael~E. Banchs, and Alexander Rudnicky. 2021.
\newblock \href {https://arxiv.org/abs/2111.02110} {Automatic evaluation and moderation of open-domain dialogue systems}.
\newblock \emph{CoRR}, abs/2111.02110.

\bibitem[{Zhang et~al.(2020)Zhang, Kishore, Wu, Weinberger, and Artzi}]{DBLP:conf/iclr/ZhangKWWA20}
Tianyi Zhang, Varsha Kishore, Felix Wu, Kilian~Q. Weinberger, and Yoav Artzi. 2020.
\newblock \href {https://openreview.net/forum?id=SkeHuCVFDr} {Bertscore: Evaluating text generation with {BERT}}.
\newblock In \emph{8th International Conference on Learning Representations, {ICLR} 2020, Addis Ababa, Ethiopia, April 26-30, 2020}. OpenReview.net.

\bibitem[{Zhao et~al.(2020)Zhao, Lala, and Kawahara}]{DBLP:conf/acl/ZhaoLK20}
Tianyu Zhao, Divesh Lala, and Tatsuya Kawahara. 2020.
\newblock \href {https://doi.org/10.18653/V1/2020.ACL-MAIN.4} {Designing precise and robust dialogue response evaluators}.
\newblock In \emph{Proceedings of the 58th Annual Meeting of the Association for Computational Linguistics, {ACL} 2020, Online, July 5-10, 2020}, pages 26--33. Association for Computational Linguistics.

\bibitem[{Zheng et~al.(2023)Zheng, Chiang, Sheng, Zhuang, Wu, Zhuang, Lin, Li, Li, Xing, Zhang, Gonzalez, and Stoica}]{DBLP:conf/nips/ZhengC00WZL0LXZ23}
Lianmin Zheng, Wei{-}Lin Chiang, Ying Sheng, Siyuan Zhuang, Zhanghao Wu, Yonghao Zhuang, Zi~Lin, Zhuohan Li, Dacheng Li, Eric~P. Xing, Hao Zhang, Joseph~E. Gonzalez, and Ion Stoica. 2023.
\newblock \href {http://papers.nips.cc/paper\_files/paper/2023/hash/91f18a1287b398d378ef22505bf41832-Abstract-Datasets\_and\_Benchmarks.html} {Judging llm-as-a-judge with mt-bench and chatbot arena}.
\newblock In \emph{Advances in Neural Information Processing Systems 36: Annual Conference on Neural Information Processing Systems 2023, NeurIPS 2023, New Orleans, LA, USA, December 10 - 16, 2023}.

\bibitem[{Zhong et~al.(2022)Zhong, Liu, Yin, Mao, Jiao, Liu, Zhu, Ji, and Han}]{DBLP:conf/emnlp/Zhong0YMJLZJH22}
Ming Zhong, Yang Liu, Da~Yin, Yuning Mao, Yizhu Jiao, Pengfei Liu, Chenguang Zhu, Heng Ji, and Jiawei Han. 2022.
\newblock \href {https://doi.org/10.18653/V1/2022.EMNLP-MAIN.131} {Towards a unified multi-dimensional evaluator for text generation}.
\newblock In \emph{Proceedings of the 2022 Conference on Empirical Methods in Natural Language Processing, {EMNLP} 2022, Abu Dhabi, United Arab Emirates, December 7-11, 2022}, pages 2023--2038. Association for Computational Linguistics.

\bibitem[{Zhou et~al.(2023)Zhou, Liu, Xu, Iyer, Sun, Mao, Ma, Efrat, Yu, Yu, Zhang, Ghosh, Lewis, Zettlemoyer, and Levy}]{DBLP:conf/nips/ZhouLX0SMMEYYZG23}
Chunting Zhou, Pengfei Liu, Puxin Xu, Srinivasan Iyer, Jiao Sun, Yuning Mao, Xuezhe Ma, Avia Efrat, Ping Yu, Lili Yu, Susan Zhang, Gargi Ghosh, Mike Lewis, Luke Zettlemoyer, and Omer Levy. 2023.
\newblock \href {http://papers.nips.cc/paper\_files/paper/2023/hash/ac662d74829e4407ce1d126477f4a03a-Abstract-Conference.html} {{LIMA:} less is more for alignment}.
\newblock In \emph{Advances in Neural Information Processing Systems 36: Annual Conference on Neural Information Processing Systems 2023, NeurIPS 2023, New Orleans, LA, USA, December 10 - 16, 2023}.

\bibitem[{Zhou et~al.(2022)Zhou, Blodgett, Trischler, III, Suleman, and Olteanu}]{DBLP:conf/naacl/ZhouBTDSO22}
Kaitlyn Zhou, Su~Lin Blodgett, Adam Trischler, Hal~Daum{\'{e}} III, Kaheer Suleman, and Alexandra Olteanu. 2022.
\newblock \href {https://doi.org/10.18653/V1/2022.NAACL-MAIN.24} {Deconstructing {NLG} evaluation: Evaluation practices, assumptions, and their implications}.
\newblock In \emph{Proceedings of the 2022 Conference of the North American Chapter of the Association for Computational Linguistics: Human Language Technologies, {NAACL} 2022, Seattle, WA, United States, July 10-15, 2022}, pages 314--324. Association for Computational Linguistics.

\bibitem[{Zhu et~al.(2023)Zhu, Wang, and Wang}]{DBLP:journals/corr/abs-2310-17631}
Lianghui Zhu, Xinggang Wang, and Xinlong Wang. 2023.
\newblock \href {https://doi.org/10.48550/ARXIV.2310.17631} {Judgelm: Fine-tuned large language models are scalable judges}.
\newblock \emph{CoRR}, abs/2310.17631.

\end{thebibliography}

\appendix

\section{Details for NLG-Eval Corpus}
\label{sec:corpus}

\begin{figure*}
\centering
\includegraphics[width=\textwidth]{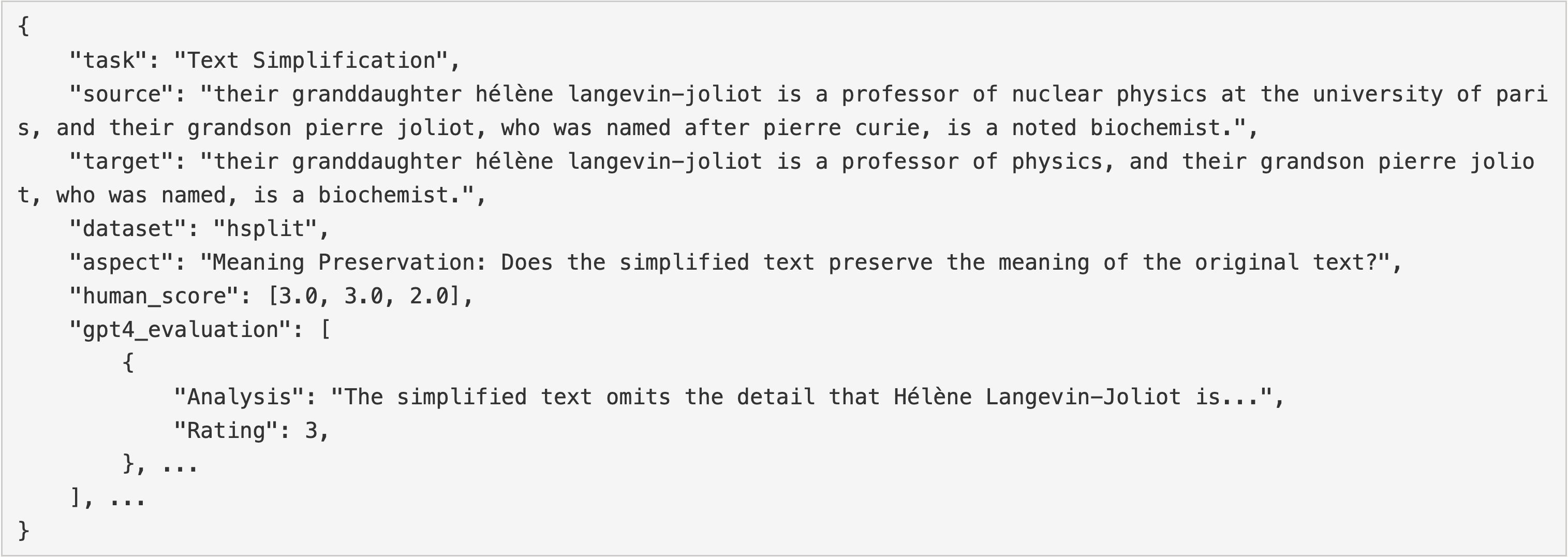}
\caption{A simplified example from the hsplit dataset for the text simplification task.}
\label{fig:dataset_example}
\end{figure*}

Our constructed NLG-Eval corpus includes evaluation datasets on nine NLG tasks, including Controllable Generation, Data to Text, Dialogue Response Generation, Grammatical Error Correction, Machine Translation, Paraphrase Generation, Story Generation, Summarization, and Text Simplification. An example from the hsplit dataset for the text simplification task is presented in Figure~\ref{fig:dataset_example} with simplified illustrations, whose situation is similar for other datasets in NLG-Eval corpus. The statistics are shown in Table~\ref{tab:statistics} and more detailed meta information is described in Table~\ref{tab:story Generation} to Table~\ref{tab:Text Simplification}. Through such data and information, we hope to facilitate related research and advance the field of NLG evaluation.

\newcolumntype{M}[1]{>{\raggedright\arraybackslash}m{#1}}
\newcolumntype{C}[1]{>{\centering\arraybackslash}m{#1}}

\begin{table*}[!htp]
\centering
\small
\renewcommand{\arraystretch}{1.25}
\begin{tabular}{M{4cm}C{1.6cm}C{1.6cm}C{1.6cm}}
\toprule
\textbf{Task} & \textbf{\#Datasets} & \textbf{\#Aspects} & \textbf{\#Samples} \\
\midrule
Controllable Generation & 4 & 8 & 11299 \\
Data to Text & 6 & 11 & 36548 \\
Dialogue Response Generation & 17 & 17 & 91111 \\
Grammatical Error Correction & 3 & 6 & 41058 \\
Machine Translation & 2 & 6 & 347504 \\
Paraphrase Generation & 2 & 3 & 18299 \\
Story Generation & 6 & 17 & 12636 \\
Summarization & 10 & 17 & 61977 \\
Text Simplification & 8 & 8 & 27026 \\
\bottomrule
\end{tabular}
\caption{The statistics of NLG-Eval corpus.}
\label{tab:statistics}
\end{table*}

\begin{table*}[!htp]
\centering
\small
\renewcommand{\arraystretch}{1}
\begin{tabular}{M{3cm}|M{0.9cm}|M{11cm}}
\toprule
\textbf{Dataset} & \textbf{Size} & \textbf{Aspect} \\

\midrule
\multirow{4}{*}{\makecell[l]{Chiang-LLM-Evaluation \\ \citep{DBLP:conf/acl/ChiangL23}}} & \multirow{4}{*}{1600} & Cohesiveness: How well do the sentences in the story fragment fit together? \\
 & & Grammaticality: How grammatically correct is the text of the story fragment? \\
 & & Likability: How enjoyable do you find the story fragment? \\
 & & Relevance: How relevant is the story fragment to the story prompt? \\
\midrule
\multirow{10}{*}{\makecell[l]{CoEval \\ \citep{DBLP:journals/corr/abs-2310-19740}}} & \multirow{10}{*}{1400} & Character Development: The characters in the generated story should be well-developed. \\
 & & Clarity: The generated story should be clear and easy to understand, with no confusing or ambiguous elements. \\
 & & Coherence: The generated story should have a logical flow and provide closure. \\
 & & Engagement: The generated story should be engaging from beginning to end. \\
 & & Grammaticality: The generated story should be grammatically correct. \\
 & & Length: The generated story should have an appropriate length according to the story requirement. \\
 & & Relevance: The generated story should be relevant to the story requirement (beginning or topic) and the daily events it aims to capture. \\
\midrule
\multirow{10}{*}{\makecell[l]{Hanna \\ \citep{DBLP:conf/coling/ChhunCSC22}}} & \multirow{10}{*}{6336} & Coherence: How much does the generated story make sense? \\
 & & Complexity: How elaborate is the generated story, involving complex concepts, realistic characters, an intricate plot, an underlying history or circumstances, or precise descriptions? \\
 & & Empathy: How well do you understand the character's emotions in the generated story (regardless of whether you agree with them)? \\
 & & Engagement: How much do you engage with the generated story? \\
 & & Relevance: How well does the generated story match the story prompt? \\
 & & Surprise: How surprising is the end of the generated story, with enough clues for a reasonable prediction? \\
\midrule
\makecell[l]{Mans\_roc \\ \citep{DBLP:conf/acl/GuanZFLDMFH20}} & 1000 & Overall Quality: The overall quality of the generated story, considering whether it has global errors like chaotic scenes (difficult to understand as a whole) and local errors, including repetitive plots (repeating similar texts), unrelated events (to the story beginning or within its own context), and conflicting logic (against common sense or with wrong causal or temporal relationship). \\
\midrule
\makecell[l]{Mans\_wp \\ \citep{DBLP:conf/acl/GuanZFLDMFH20}} & 1000 & Overall Quality: The overall quality of the generated story, considering whether it has global errors like chaotic scenes (difficult to understand as a whole) and local errors, including repetitive plots (repeating similar texts), unrelated events (to the story prompt or within its own context), and conflicting logic (against common sense or with wrong causal or temporal relationship). \\
\midrule
\multirow{5}{*}{\makecell[l]{nextchapter \\ \citep{DBLP:conf/inlg/XieCL23}}} & \multirow{5}{*}{1300} & Coherence: How well do the sentences in the generated story fit together? \\
 & & Fluency: How grammatically correct is the text of the generated story? \\
 & & Interestingness: How enjoyable do you find the generated story? \\
 & & Logicality: How much does the generated story obey commonsense? \\
 & & Relatedness: How relevant is the generated story to the story prompt? \\
\bottomrule
\end{tabular}
\caption{Datasets for Story Generation task.}
\label{tab:story Generation}
\end{table*}

\begin{table*}[!htp]
\small
\centering
\renewcommand{\arraystretch}{1}
\begin{tabular}{M{3cm}|M{0.9cm}|M{11cm}}
\toprule
\textbf{Dataset} & \textbf{Size} & \textbf{Aspect} \\

\midrule
\multirow{5}{*}{\makecell[l]{CTRLEval \\ \citep{DBLP:conf/acl/KeZLLZZH22}}} & \multirow{5}{*}{3960} & Attribute Relevance: Measure whether the generated text satisfies the attribute label. \\
 & & Coherence: Measure whether the sentences in the generated text are semantically relevant to compose a coherent body, which reflects the quality of the generated text itself. \\
 & & Consistency: Evaluate whether the generated text is consistent to the content prefix. \\
\midrule
\makecell[l]{FUDGE \\ \citep{DBLP:conf/naacl/YangK21}} & \raisebox{-3pt}{2088} & \raisebox{-3pt}{Fluency: Is the generated text fluent, i.e., well-written and grammatical?} \\
\midrule
\raisebox{3pt}{\makecell[l]{PPLM \\ \citep{DBLP:conf/iclr/DathathriMLHFMY20}}} & 
\raisebox{-1pt}{3251} & Fluency: Whether the generated text has no grammatical errors, formatting problems, or obviously ungrammatical issues (e.g., fragments, missing components) that make the text difficult to read? \\
\midrule
\multirow{7}{*}{\makecell[l]{InstruSum \\ \citep{DBLP:journals/corr/abs-2311-09184}}} & \multirow{7}{*}{2000} & Factual Consistency: Is the summary consistent with the facts presented in the article, without contradicting or misrepresenting any information? \\
 & & Irrelevant Information: Does the summary include any information that is not relevant to the summary requirement? \\
 & & Missing Information: Does the summary omit any crucial information from the article concerning the summary requirement? \\
 & & Overall Quality: Assess the overall quality of the summary in relation to the summary requirement. \\

\bottomrule
\end{tabular}
\caption{Datasets for Controllable Generation task.}
\label{tab:Controllable Generation}
\end{table*}

\begin{table*}[!htp]
\centering
\small
\renewcommand{\arraystretch}{1}
\begin{tabular}{M{3cm}|M{0.9cm}|M{11cm}}
\toprule
\textbf{Dataset} & \textbf{Size} & \textbf{Aspect} \\

\midrule
\multirow{3}{*}{\makecell[l]{E2E NLG \\ \citep{DBLP:journals/csl/DusekNR20}}} & \multirow{3}{*}{6300} & Naturalness: Could the utterance have been produced by a native speaker? \\
 & & Overall Quality: How is the overall quality of the utterance in terms of its grammatical correctness, fluency, adequacy and other important factors? \\
\midrule
\multirow{4}{*}{\makecell[l]{INLG16 \\ \citep{DBLP:conf/inlg/NovikovaLR16}}} & \multirow{4}{*}{3726} & Informativeness: Is this utterance informative? (i.e. do you think it provides enough useful information from the data?) \\
 & & Naturalness: Is this utterance natural? (e.g. could it have been produced by a native speaker?) \\
 & & Phrasing: Is this utterance well phrased? (i.e. do you like how it is expressed?) \\
\midrule
\multirow{4}{*}{\makecell[l]{RankMe \\ \citep{DBLP:conf/naacl/NovikovaDR18}}} & \multirow{4}{*}{900} & Informativeness (= adequacy): Does the utterance provide all the useful information from the meaning representation? \\
 & & Naturalness (= fluency): Could the utterance have been produced by a native speaker? \\
 & & Overall Quality: How do you judge the overall quality of the utterance in terms of its grammatical correctness, fluency, adequacy and other important factors? \\
\midrule
\multirow{4}{*}{\makecell[l]{SFRES\_SFHOT \\ \citep{DBLP:conf/emnlp/WenGMSVY15}}} & \multirow{4}{*}{6168} & Informativeness: Does the utterance provide all the useful information from the meaning representation? \\
 & & Naturalness: Could the utterance have been produced by a native speaker? \\
 & & Overall Quality: How do you judge the overall quality of the utterance in terms of its grammatical correctness and fluency? \\
\midrule
\multirow{3}{*}{\makecell[l]{webnlg\_2017 \\ \citep{DBLP:conf/acl/GardentSNP17}}} & \multirow{3}{*}{5214} & Fluency: Does the text sound fluent and natural? \\
 & & Grammaticality: Is the text grammatical (no spelling or grammatical errors)? \\
 & & Semantic Adequacy: Does the text correctly represent the meaning in the data? \\
\midrule
\multirow{8}{*}{\makecell[l]{\parbox{2.8cm}{webnlg\_2020 \\ \citep{castro-ferreira-etal-2020-2020}}}} & \multirow{8}{*}{14240} & Correctness: When describing predicates which are found in the data, does the text mention the correct objects and adequately introduce the subject for this specific predicate? \\
 & & Data Coverage: Does the text include descriptions of all predicates presented in the data? \\
 & & Fluency: Is it possible to say that the text progresses naturally, forms a coherent whole and it is easy to understand the text? \\
 & & Relevance: Does the text describe only such predicates (with related subjects and objects), which are found in the data? \\
 & & Text Structure: Is the text grammatical, well-structured, written in acceptable English language? \\

\bottomrule
\end{tabular}
\caption{Datasets for Data to Text task.}
\label{tab:Data-to-Text}
\end{table*}

\begin{table*}[!htp]
\centering
\small
\renewcommand{\arraystretch}{1}
\begin{tabular}{M{3cm}|M{0.9cm}|M{11cm}}
\toprule
\textbf{Dataset} & \textbf{Size} & \textbf{Aspect} \\

\midrule
\makecell[l]{convai2-grade \\ \citep{DBLP:conf/emnlp/HuangYQLL20}} & \raisebox{-2.5pt}{600} & \raisebox{-2.5pt}{\parbox{11cm}{Coherence: The response should be coherent with the dialogue context, maintaining a good logical flow.}} \\
\midrule
\makecell[l]{dailydialog-grade \\ \citep{DBLP:conf/emnlp/HuangYQLL20}} & \raisebox{-2.75pt}{300} & \raisebox{-2.5pt}{\parbox{11cm}{Coherence: The response should be coherent with the dialogue context, maintaining a good logical flow.}} \\
\midrule
\makecell[l]{dailydialog-gupta \\ \citep{DBLP:conf/sigdial/GuptaMZPEB19}} & \raisebox{-2.5pt}{500} & \raisebox{-2.5pt}{\parbox{11cm}{Appropriateness: Whether the response is appropriate given the dialogue context in grammar, topic, and logic?}} \\
\midrule
\multirow{6}{*}{\makecell[l]{dailydialog-zhao \\ \citep{DBLP:conf/acl/ZhaoLK20}}} & \multirow{6}{*}{3600} & Appropriateness: The response should be appropriate given the preceding dialogue. \\
 & & Content Richness: The response should be informative, with long sentences including multiple entities and conceptual or emotional words. \\
 & & Grammatical Correctness: The response should be free of grammatical and semantic errors. \\
 & & Relevance: The response should be on-topic with the immediate dialogue history. \\
\midrule
\multirow{6}{*}{\makecell[l]{DialogADV \\ \citep{liu2023evaluate}}} & \multirow{6}{*}{16416} & Coherence: The logical and semantic coherence between the response and dialogue history (previous context). \\
 & & Consistency: The logical and factual consistency between the response and dialogue history (previous context), facts also include external commonsense knowledge. \\
 & & Fluency: The fluency and grammatical correctness of the response. \\
 & & Relevance: The degree to response is connected or relevant to a particular topic, question, or situation of the dialogue history (previous context). \\
 \midrule
\multirow{6}{*}{\makecell[l]{dstc10-persona\_clean \\ \citep{DBLP:journals/corr/abs-2111-02110}}} & \multirow{6}{*}{19316} & Appropriateness: The response should be appropriate given the preceding dialogue. \\
 & & Content Richness: The response should be informative, with long sentences including multiple entities and conceptual or emotional words. \\
 & & Grammatical Correctness: The response should be free of grammatical and semantic errors. \\
 & & Relevance: The response should be on-topic with the immediate dialogue history. \\
 \midrule
\multirow{6}{*}{\makecell[l]{dstc10-topical\_clean \\ \citep{DBLP:journals/corr/abs-2111-02110}}} & \multirow{6}{*}{18000} & Appropriateness: The response should be appropriate given the preceding dialogue. \\
 & & Content Richness: The response should be informative, with long sentences including multiple entities and conceptual or emotional words. \\
 & & Grammatical Correctness: The response should be free of grammatical and semantic errors. \\
 & & Relevance: The response should be on-topic with the immediate dialogue history. \\
 \midrule
\makecell[l]{empathetic-grade \\ \citep{DBLP:conf/emnlp/HuangYQLL20}} & \raisebox{-2.75pt}{300} & \raisebox{-2.5pt}{\parbox{11cm}{Coherence: The response should be coherent with the dialogue context, maintaining a good logical flow.}} \\
\midrule
\makecell[l]{esl \\ \citep{DBLP:journals/corr/abs-2010-12741}} & \raisebox{-2.5pt}{1242} & \raisebox{-2.5pt}{\parbox{11cm}{Appropriateness: The response should be appropriate given the dialogue context, in grammar, topic, and logic.}} \\
\midrule
\multirow{13}{*}{\makecell[l]{\parbox{2.8cm}{fed-turn \\ \citep{DBLP:conf/sigdial/MehriE20}}}} & \multirow{13}{*}{3375} & Correctness: Is the response correct or was there a misunderstanding of the conversation? \\
 & & Engagingness: Is the response engaging to user and fulfill the particular conversational goals implied by the user? \\
 & & Fluency: Is the response fluently written and free of grammatical and semantic errors? \\
 & & Interestingness: To the average person, is the response interesting? \\
 & & Overall Quality: What is the overall impression of the response, and quality of and satisfaction with the conversation?  \\
 & & Relevance: Is the response relevant to and on-topic with the conversation history? \\
 & & Semantical Appropriateness: Is the response semantically appropriate given the conversation history? \\
 & & Specificity: Does the response produce unique and non-generic information that is specific to the conversation history?  \\
 & & Understandability: Is the response understandable? \\
 \midrule
\multirow{3}{*}{\makecell[l]{holistic dialogue \\ \citep{DBLP:conf/acl/PangNHZLT20}}} & \multirow{3}{*}{400} & Coherence: Measures the meaningfulness of the response within the context of prior query. \\
 & & Fluency: Measures the quality of phrasing of the response relative to a human native speaker. \\
\midrule
\multirow{2}{*}{\makecell[l]{humod \\ \citep{merdivan2020human}}} & \multirow{2}{*}{19000} & Fluency: The response should be written naturally and free of grammatical and semantic errors. \\
 & & Relevance: The Response should be on-topic with the immediate dialogue history. \\
 \midrule
\raisebox{-1.5pt}{\makecell[l]{jsalt \\ \citep{DBLP:conf/iwsds/Kong-VegaSWD18}}} & \raisebox{-3.5pt}{741} & \raisebox{-2pt}{\parbox{11cm}{Appropriateness: Measure how well the response is semantically and pragmatically valid given the previous recent dialogue history.}} \\
\bottomrule
\end{tabular}
\caption{Datasets for Dialogue Response Generation task. (Part 1)}
\label{tab:Dialogue Response Generation (Part 1)}
\end{table*}

\begin{table*}[!htp]
\centering
\small
\renewcommand{\arraystretch}{1}
\begin{tabular}{M{3cm}|M{0.9cm}|M{11cm}}
\toprule
\textbf{Dataset} & \textbf{Size} & \textbf{Aspect} \\

\midrule
\raisebox{3pt}{\makecell[l]{ncm \\ \citep{DBLP:conf/naacl/SedocIKTUC19}}} & \raisebox{-1.5pt}{2461} & Appropriateness: Whether the response is appropriate given the dialogue history, in grammar, topic, and logic? \\
\midrule
\multirow{10}{*}{\makecell[l]{persona-usr \\ \citep{DBLP:conf/naacl/SedocIKTUC19}}} & \multirow{10}{*}{1800} & Context Maintenance: Does the response serve as a valid continuation of the dialogue context (conversation history)? \\
 & & Interestingness: Is the response dull or interesting? \\
 & & Knowledge Use: Given the fact that the response is conditioned on, how well does the response use that fact? \\
 & & Naturalness: Does the response seem to be something that a person would naturally say? \\
 & & Overall Quality: What is the overall impression of this utterance? Please consider whether the response is understandable and natural, and how well it maintains context and uses knowledge. \\
 & & Understandability: Is the response understandable given the previous dialogue context? (Not if its on topic, but for example if it uses pronouns they should make sense) \\
 \midrule
\makecell[l]{persona-zhao \\ \citep{DBLP:conf/acl/ZhaoLK20}} & \raisebox{-3pt}{900} & \raisebox{-3pt}{Appropriateness: The response should be appropriate given the preceding dialogue.} \\
\midrule
\multirow{10}{*}{\makecell[l]{topical-usr \\ \citep{DBLP:conf/naacl/SedocIKTUC19}}} & \multirow{10}{*}{2160} & Context Maintenance: Does the response serve as a valid continuation of the dialogue context (conversation history)? \\
 & & Interestingness: Is the response dull or interesting? \\
 & & Knowledge Use: Given the fact that the response is conditioned on, how well does the response use that fact? \\
 & & Naturalness: Does the response seem to be something that a person would naturally say? \\
 & & Overall Quality: What is the overall impression of this utterance? Please consider whether the response is understandable and natural, and how well it maintains context and uses knowledge. \\
 & & Understandability: Is the response understandable given the previous dialogue context? (Not if its on topic, but for example if it uses pronouns they should make sense) \\
\bottomrule
\end{tabular}
\caption{Datasets for Dialogue Response Generation task. (Part 2)}
\label{tab:Dialogue Response Generation (Part 2)}
\end{table*}

\begin{table*}[!htp]
\small
\centering
\renewcommand{\arraystretch}{1}
\begin{tabular}{M{3cm}|M{0.9cm}|M{11cm}}
\toprule
\textbf{Dataset} & \textbf{Size} & \textbf{Aspect} \\

\midrule
\makecell[l]{GMEG \\ \citep{DBLP:journals/tacl/NapolesNT19}} & \raisebox{-2pt}{27195} & \raisebox{-2pt}{\parbox{11cm}{Overall Quality: Whether the corrected text is perfect, namely grammatical and not garbled?}} \\
\midrule
\multirow{7}{*}{\makecell[l]{protagolabs \\ \citep{DBLP:conf/emnlp/SottanaLZY23}}} & \multirow{7}{*}{1200} & Grammaticality: The quality of the correction and the extent to which errors are left in the corrected text, regardless of whether they are present in the original text or they are newly introduced errors in the supposed corrected version. \\
 & & Over-correction: Assess whether the correction avoids unnecessary syntax changes or being unnecessarily verbose, since there can be multiple ways to correct a text. The best correction should be done with the minimum number of edits. \\
 & & Semantics: Whether the meaning of the original text is preserved following the grammatical error correction? NOTE: You should penalize corrections which change the meaning unnecessarily. \\
\midrule
\multirow{5}{*}{\makecell[l]{TMU-GFM \\ \citep{DBLP:conf/coling/YoshimuraKKK20}}} & \multirow{5}{*}{12663} & Fluency: How natural does the corrected text sound for native speakers? \\
 & & Grammaticality: The grammatical correctness of the corrected text and how comprehensible it is. \\
 & & Meaning Preservation: The extent to which the meaning of the original text is preserved in the corrected text. \\

\bottomrule
\end{tabular}
\caption{Datasets for Grammatical Error Correction task.}
\label{tab:Grammatical Error Correction}
\end{table*}

\begin{table*}[!htp]
\centering
\small
\renewcommand{\arraystretch}{1}
\begin{tabular}{M{3cm}|M{0.9cm}|M{11cm}}
\toprule
\textbf{Dataset} & \textbf{Size} & \textbf{Aspect} \\
\midrule
\multirow{3}{*}{\makecell[l]{parabank \\ \citep{DBLP:conf/aaai/HuRPD19}}} & \multirow{3}{*}{11140} & Fluency: Whether the paraphrase is meaningful and grammatical? \\
 & & Semantic Similarity: Whether the paraphrase maintains similar semantics to the original text? \\
\midrule
\raisebox{3pt}{\makecell[l]{twitter para \\ \citep{DBLP:conf/emnlp/ShenLJ022}}} & \raisebox{-1.5pt}{7159} & Overall Quality: The paraphrase should not only maintain similar semantics to the original text, but also possess lexical or syntactic differences from the original text, with fluent and coherent content. \\
\bottomrule
\end{tabular}
\caption{Datasets for Paraphrase Generation task.}
\label{tab:Paraphrase Generation}
\end{table*}

\begin{table*}[!htp]
\centering
\small
\renewcommand{\arraystretch}{1}
\begin{tabular}{M{3cm}|M{0.9cm}|M{11cm}}
\toprule
\textbf{Dataset} & \textbf{Size} & \textbf{Aspect} \\

\midrule
\multirow{13}{*}{\makecell[l]{WMT\_zhen \\ \citep{DBLP:journals/tacl/FreitagFGRTM21}}} & \multirow{13}{*}{346504} & Accuracy: The translation should accurately represent the source text, not including information not present in the source text or missing content from the source text. \\
 & & Fluency: The translation should not have incorrect punctuation and spelling, problems with grammar, internal inconsistency, or garbled characters due to incorrect encoding. \\
 & & Locale Convention: The translation should not have the wrong formats for addresses, currency, dates, names, telephone numbers, and time expressions. \\
 & & Overall Quality: The overall quality of the translation, including accuracy (e.g., mistranslation, omission, addition), fluency (e.g., grammar, spelling, punctuation, inconsistency), locale convention (e.g., format for names, currency, address), terminology (e.g., inappropriate or inconsistent usage), and style (e.g., stylistic problems). \\
 & & Style: Does the translation have no stylistic problems? \\
 & & Terminology: Whether the terminology of the translation is standard, appropriate for the context, and used consistently? \\
\midrule
\makecell[l]{HumanMT \\ \citep{DBLP:conf/acl/RiezlerKU18}} & 1000 & Overall Quality: The overall quality of the translation, including accuracy (e.g., mistranslation, omission, addition), fluency (e.g., grammar, spelling, punctuation, inconsistency), locale convention (e.g., format for names, currency, address), terminology (e.g., inappropriate or inconsistent usage), and style (e.g., stylistic problems). \\
\bottomrule
\end{tabular}
\caption{Datasets for Machine Translation task.}
\label{tab:Translation}
\end{table*}

\begin{table*}[!htp]
\centering
\small
\renewcommand{\arraystretch}{1}
\begin{tabular}{M{3cm}|M{0.9cm}|M{11cm}}
\toprule
\textbf{Dataset} & \textbf{Size} & \textbf{Aspect} \\
\midrule
\multirow{8}{*}{\makecell[l]{DialSummEval \\ \citep{DBLP:conf/naacl/Gao022}}} & \multirow{8}{*}{5600} & Coherence: Measure the quality of all sentences of the summary collectively, to fit together and sound naturally. Consider the quality of the summary as a whole. \\
 & & Consistency: Measure whether the facts in the summary are consistent with the facts in the dialogue. Consider whether the summary does reproduce all facts accurately and does not make up untrue information. \\
 & & Fluency: Measure the quality of individual sentences of the summary, whether they are well-written and grammatically correct. Consider the quality of individual sentences. \\
 & & Relevance: Measure how well the summary captures the key points of the dialogue. Consider whether all and only the important aspects are contained in the summary. \\
\midrule
\makecell[l]{frank \\ \citep{DBLP:conf/naacl/PagnoniBT21}} & \raisebox{-3pt}{2246} & \raisebox{-3pt}{\parbox{11cm}{Factuality: Measure whether the facts in the summary are correct according to the article.}} \\
\midrule
\multirow{5}{*}{\makecell[l]{Newsroom \\ \citep{DBLP:conf/naacl/GruskyNA18}}} & \multirow{5}{*}{1680} & Coherence: Do phrases and sentences of the summary fit together and make sense collectively? \\
 & & Fluency: Are the individual sentences of the summary well-written and grammatical? \\
 & & Informativeness: How well does the summary capture the key points of the article? \\
 & & Relevance: Are the details provided by the summary consistent with details in the article? \\
\midrule
\multirow{16}{*}{\makecell[l]{OpenAI \\ \citep{DBLP:journals/corr/abs-2009-01325}}} & \multirow{16}{*}{34197} & Accuracy: Does the factual information in the summary accurately match the post? A summary is accurate if it doesn't say things that aren't in the post, it doesn't mix up people, and generally is not misleading. \\
 & & Coherence: How coherent is the summary on its own? A summary is coherent if, when read by itself, it's easy to understand and free of English errors. A summary is not coherent if it's difficult to understand what the summary is trying to say. Generally, it's more important that the summary is understandable than it being free of grammar errors. \\
 & & Coverage: How well does the summary cover the important information in the post? A summary has good coverage if it mentions the main information from the post that's important to understand the situation described in the post. A summary has poor coverage if someone reading only the summary would be missing several important pieces of information about the situation in the post. A summary with good coverage should also match the purpose of the original post (e.g. to ask for advice). \\
 & & Overall Quality: How good is the summary overall at representing the post? This can encompass coherence (how coherent is the summary on its own), accuracy (does the factual information in the summary accurately match the post), and coverage (how well does the summary cover the important information in the post) of the summary, as well as other important aspects. \\
\midrule
\raisebox{-6pt}{\makecell[l]{QAGS \\ \citep{DBLP:conf/acl/WangCL20}}} & \raisebox{-6pt}{474} & \raisebox{-2pt}{\parbox{11cm}{Factual Consistency: Is the summary factually consistent with the article? Non-grammatical sentences should be considered not consistent and copies of article sentences should be considered consistent.}} \\
\bottomrule
\end{tabular}
\caption{Datasets for Summarization task. (Part 1)}
\label{tab:Summarization (part 1)}
\end{table*}

\begin{table*}[!htp]
\centering
\small
\renewcommand{\arraystretch}{1}
\begin{tabular}{M{3cm}|M{0.9cm}|M{11cm}}
\toprule
\textbf{Dataset} & \textbf{Size} & \textbf{Aspect} \\
 \midrule
\multirow{9}{*}{\makecell[l]{OpinSummEval \\ \citep{DBLP:journals/corr/abs-2310-18122}}} & \multirow{9}{*}{5600} & Aspect Relevance: Measure whether the mainly discussed aspects in the reviews are covered exactly by the summary. It focuses on whether the summary correctly reflects the mainly discussed aspects in the reviews. \\
 & & Readability: Measure whether the summary is fluent and informative. It focuses on whether the summary is well-written and valuable. \\
 & & Self-coherence: Measure whether the summary is consistent within itself in terms of sentiments and aspects. It focuses on whether the summary is coherent and does not reflect conflicting opinions. \\
 & & Sentiment Consistency: Measure whether the summary is consistent with the reviews in terms of sentiments for each aspect. It focuses on whether the summary aspect-wisely captures the main sentiment in the reviews. \\
\midrule
\makecell[l]{PolyTope \\ \citep{DBLP:conf/emnlp/HuangCYBWXZ20}} & 1268 & Overall Quality: The overall quality of the summary, including accuracy and fluency. Accuracy-related issues refer to the extent to which the summary does not match the article, including unnecessary snippets, missing key points, content unfaithful to the article, content not present in the article and factually incorrect, and statements that contradict the article in attitudes (e.g. from positive to negative). Fluency-related issues refer to the linguistic quality of the summary, which is independent of the relationship between the article and the summary, including unnecessary repetition, problems in the word form, and problems in word order. \\
\midrule
\multirow{8}{*}{\makecell[l]{protagolabs \\ \citep{DBLP:conf/emnlp/SottanaLZY23}}} & \multirow{8}{*}{1600} & Coherence: Measure the quality of all sentences of the summary collectively, to fit together and sound naturally. Consider the quality of the summary as a whole. \\
 & & Consistency: Measure whether the facts in the summary are consistent with the facts in the article. Consider whether the summary does reproduce all facts accurately and does not make up untrue information. \\
 & & Fluency: Measure the quality of individual sentences of the summary, whether they are well-written and grammatically correct. Consider the quality of individual sentences. \\
 & & Relevance: Measure how well the summary captures the key points of the article. Consider whether all and only the important aspects are contained in the summary. \\
\midrule
\multirow{8}{*}{\makecell[l]{SummEval \\ \citep{DBLP:journals/tacl/FabbriKMXSR21}}} & \multirow{8}{*}{6400} & Coherence: Measure the quality of all sentences of the summary collectively, to fit together and sound naturally. Consider the quality of the summary as a whole. \\
 & & Consistency: Measure whether the facts in the summary are consistent with the facts in the article. Consider whether the summary does reproduce all facts accurately and does not make up untrue information. \\
 & & Fluency: Measure the quality of individual sentences of the summary, whether they are well-written and grammatically correct. Consider the quality of individual sentences. \\
 & & Relevance: Measure how well the summary captures the key points of the article. Consider whether all and only the important aspects are contained in the summary. \\
\midrule
\multirow{23}{*}{\makecell[l]{SummEval-OP \\ \citep{DBLP:journals/corr/abs-2402-11683}}} & \multirow{23}{*}{2912} & Aspect Coverage: The summary should cover all the aspects that are majorly being discussed in the reviews. The summary should be penalized if it misses out on an aspect that was majorly discussed in the reviews and awarded if it covers all. \\
 & & Coherence: Measure the collective quality of all sentences. The summary should be well-structured and well-organized. The summary should not just be a heap of related information, but should build from sentences to a coherent body of information. \\
 & & Faithfulness: Every piece of information mentioned in the summary should be verifiable/supported/inferred from the reviews only. The summary should be penalized if any piece of information is not verifiable/supported/inferred from the reviews or if the summary overgeneralizes something. \\
 & & Fluency: Measure the quality of the summary in terms of grammar, spelling, punctuation, capitalization, word choice, and sentence structure and should contain no errors. The summary should be easy to read, follow, comprehend and should contain no errors.\\
 & & Relevance: The summary should not contain opinions that are either not consensus or important. The summary should include only important opinions from the reviews. The summary should be penalized if it contains redundancies and unimportant information. \\
 & & Sentiment Consistency: All the aspects discussed in the summary should accurately reflect the consensus sentiment of the corresponding aspects from the reviews. The summary should be penalized if it does not cover accurately the sentiment regarding any aspect within the summary. \\
 & & Specificity: The summary should avoid containing generic opinions. All the opinions within the summary should contain detailed and specific information about the consensus opinions. The summary should be penalized for missing out details and should be awarded if they are specific. \\
\bottomrule
\end{tabular}
\caption{Datasets for Summarization task. (Part 2)}
\label{tab:Summarization (part 2)}
\end{table*}

\begin{table*}[!htp]
\centering
\small
\renewcommand{\arraystretch}{1}
\begin{tabular}{M{3cm}|M{0.9cm}|M{11cm}}
\toprule
\textbf{Dataset} & \textbf{Size} & \textbf{Aspect} \\

\midrule
\multirow{5}{*}{\makecell[l]{\parbox{3cm}{ASSET \\ \citep{DBLP:conf/acl/Alva-ManchegoMB20}}}} & \multirow{5}{*}{300} & Adequacy (or Meaning Preservation): The simplified sentence should adequately express the meaning of the original sentence, perhaps omitting the least important information. \\
 & & Fluency (or Grammaticality): The simplified sentence should be fluent, and there are no grammatical errors. \\
 & & Simplicity: The simplified sentence should be easier to understand than the original sentence. \\
\midrule
\multirow{3}{*}{\makecell[l]{Fusion \\ \citep{schwarzer-etal-2021-improving}}} & \multirow{3}{*}{10490} & Adequacy: To which degree does the simplified sentence retain the meaning of the original sentence? \\
 & & Simplicity: To which degree is the simplified sentence simpler than the original sentence? \\
\midrule
\multirow{6}{*}{\makecell[l]{HSplit \\ \citep{DBLP:conf/acl/AbendRS18}}} & \multirow{6}{*}{7560} & Grammaticality: Is the simplified text fluent and grammatical? \\
 & & Meaning Preservation: Does the simplified text preserve the meaning of the original text? \\
 & & Simplicity: Is the simplified text simpler than the original text? \\
 & & Structural Simplicity: Is the simplified text simpler than the original text, ignoring the complexity of the words? \\
\midrule
\multirow{3}{*}{\makecell[l]{Human-likert \\ \citep{DBLP:journals/corr/abs-2104-07560}}} & \multirow{3}{*}{336} & Fluency: How fluent is the simplified text? \\
 & & Meaning Preservation: How well does the simplified text express the original meaning? \\
 & & Simplicity: To what extent is the simplified text easier to read and understand? \\
\midrule
\multirow{6}{*}{\makecell[l]{LENS \\ \citep{DBLP:conf/acl/MaddelaDHX23}}} & \multirow{6}{*}{3840} & Overall Quality: The simplified sentence should be fully simplified, entirely fluent, and preserve the core meaning of the original sentence. \\
 & & Adequacy: The simplified sentence should adequately express the meaning of the original sentence, perhaps omitting the least important information. \\
 & & Fluency: The simplified sentence should be fluent, and there are no grammatical errors. \\
 & & Simplicity: The simplified sentence should be easier to understand than the original sentence. \\
\midrule
\multirow{6}{*}{\makecell[l]{\parbox{3cm}{metaeval \\ \citep{DBLP:journals/coling/Alva-ManchegoSS21}}}} & \multirow{6}{*}{1800} & Adequacy (or Meaning Preservation): Judge by looking at both the original and simplified texts, and judge whether or not the changes made preserve the original meaning. \\
 & & Fluency (or Grammaticality): Judge by looking solely at the simplified text. Mainly consider the grammatical and/or spelling errors, but also 'how well' (or natural) the text reads. Do not take capitalization into consideration. \\
 & & Simplicity: Judge by looking at both the original and simplified texts, and judge whether or not the changes made the simplified text easier to understand than the original text. \\
\midrule
\multirow{4}{*}{\makecell[l]{protagolabs \\ \citep{DBLP:conf/emnlp/SottanaLZY23}}} & \multirow{4}{*}{1200} & Fluency (or Grammaticality): Whether the simplified text remains grammatical and understandable? \\
 & & Semantics (or Adequacy): Whether the meaning is preserved further to the simplification? \\
 & & Simplicity: Whether the simplified text is simpler than the original text? \\
\midrule
\multirow{5}{*}{\makecell[l]{SAMSA \\ \citep{DBLP:conf/naacl/SulemAR18}}} & \multirow{5}{*}{1500} & Grammaticality: Is the simplified text grammatical? \\
 & & Meaning Preservation: Does the simplified text add information or remove important information, compared to the original text? \\
 & & Structural Simplicity: Is the simplified text simpler than the original text, ignoring the complexity of the words? \\

\bottomrule
\end{tabular}
\caption{Datasets for Text Simplification task.}
\label{tab:Text Simplification}
\end{table*}

\section{Settings and Prompts for GPT-4 Annotations}
\label{sec:prompt}

We follow the suggestions of~\citet{DBLP:journals/corr/abs-2312-16171} to design the instructions and prompts for GPT-4 to annotate NLG evaluation tasks, as shown in Table~\ref{tab:prompt}. Those for re-evaluating evaluations are shown in Table~\ref{tab:prompt_c1} and Table~\ref{tab:prompt_c2}, where the task descriptions are appropriately modified to prevent confusion for GPT-4.

\begin{table*}
\centering\small
\begin{tabular}{p{14cm}}
\toprule
\textbf{Prompts and Instructions} \\
\midrule
  
\#\#\#Instruction\#\#\# \\
Please act as an impartial and helpful evaluator for natural language generation (NLG), and the audience is an expert in the field. \\
Your task is to evaluate the quality of \{task\} strictly based on the given evaluation criterion. \\
Begin the evaluation by providing your analysis concisely and accurately, and then on the next line, start with "Rating:" followed by your rating on a Likert scale from 1 to 5 (higher means better). \\
You MUST keep to the strict boundaries of the evaluation criterion and focus solely on the issues and errors involved; otherwise, you will be penalized. \\
Make sure you read and understand these instructions, as well as the following evaluation criterion and example content, carefully. \\
\\
\#\#\#Evaluation Criterion\#\#\# \\
\{aspect\} \\
\\
\#\#\#Example\#\#\# \\
\{source\_des\}: \\
\{source\} \\
\\
\{target\_des\}: \\
\{target\} \\
\\
\#\#\#Your Evaluation\#\#\# \\
\\
\bottomrule
\end{tabular}
\caption{Prompts and instructions used for evaluating NLG tasks.}
\label{tab:prompt}
\end{table*}

\begin{table*}
\centering\small
\begin{tabular}{p{14cm}}
\toprule
\textbf{Prompts and Instructions} \\
\midrule
  
\#\#\#Instruction\#\#\#\\
You are a professional and helpful evaluator for natural language generation (NLG).\\
You will be given an example of Review Generation task, which includes an aspect description and a review based on it.\\
Your task is to evaluate the quality of the review strictly based on the given evaluation criterion.\\
Your evaluation MUST begin with the accurate analysis, followed by 'Rating:' and then include the corresponding evaluation rating.\\
Your rating MUST be an integer ranging from 1 to 5, following a five-point Likert scale. (higher means better)\\
You MUST keep to the strict boundaries of the given evaluation criterion and focus ONLY on the issues and errors involved; otherwise, you will be penalized.\\
Make sure you read and understand these instructions, as well as the following evaluation criterion and example content, carefully.\\
\\
\#\#\#Evaluation Criterion\#\#\#\\
Aspect Alignment: Is the review strictly aligned with and solely based on the corresponding aspect description, without mentioning any other points out of scope?\\
\\
\#\#\#Example\#\#\#\\
Aspect Description:\\
\{aspect\}\\
\\
Review:\\
\{analysis\}\\
\\
\#\#\#Your Evaluation\#\#\#\\
\\
\bottomrule
\end{tabular}
\caption{Prompts and instructions used for re-evaluating evaluations based on the alignment between evaluation analyses and aspects.}
\label{tab:prompt_c1}
\end{table*}

\begin{table*}
\centering\small
\begin{tabular}{p{14cm}}
\toprule
\textbf{Prompts and Instructions} \\
\midrule
  
\#\#\#Instruction\#\#\# \\
You are a professional and helpful evaluator for natural language generation (NLG).\\
You will be given an example of Review Generation task, which includes a review of the \{target\_des\} and a polarity.\\
Your task is to evaluate the quality of the review strictly based on the given evaluation criterion.\\
Your evaluation MUST begin with the accurate analysis, followed by 'Rating:' and then include the corresponding evaluation rating.\\
Your rating MUST be an integer ranging from 1 to 5, following a five-point Likert scale. (higher means better)\\
You MUST keep to the strict boundaries of the given evaluation criterion and focus ONLY on the issues and errors involved; otherwise, you will be penalized.\\
Make sure you read and understand these instructions, as well as the following evaluation criterion and example content, carefully.\\
\\
\#\#\#Evaluation Criterion\#\#\#\\
Polarity Consistency: Is the polarity of the review towards the \{target\_des\} consistent with the given polarity (including negative, slightly negative, neutral, slightly positive, and positive)?\\
\\
\#\#\#Example\#\#\#\\
Polarity:\\
\{rating\_to\_polarity\} \\
\\
Review:\\
\{analysis\}\\
\\
\#\#\#Your Evaluation\#\#\#\\
\\
\bottomrule
\end{tabular}
\caption{Prompts and instructions used for re-evaluating evaluations based on the alignment between evaluation analyses and ratings.}
\label{tab:prompt_c2}
\end{table*}

\section{Training Details}
\label{sec:training}

The sizes of training data used in our supervised fine-tuning and preference alignment are 67,180 and 10,000, respectively. In the construction of the latter, we prioritize selecting preference pairs with larger differences in evaluation ratings, as intuitively, their preference relations are more reliable. In addition, following the construction method of the training data, we built an additional validation set based on the NLG Eval corpus, comprising a total of 2,479 samples, to find the most suitable training hyperparameters. During supervised fine-tuning, the learning rate is 1e-5, the batch size is 128, the training epoch is 3, and the optimizer is AdamW with the zero weight decay. And in preference alignment, the learning rate is 3e-6, the batch size is 128, the training epoch is 1, $\alpha$ is 1.0, and the optimizer is AdamW. Our experiments utilize 8 A100 GPUs during training and inference.

\section{Complete Results of Six NLG Tasks}
\label{sec:6NLG}

We display the complete results of our Themis and other models on six NLG tasks respectively in Table~\ref{tab:Summarization} to Table~\ref{tab:Machine Translation}.

\begin{table*}
\centering
\small
\renewcommand{\arraystretch}{1.5}
\begin{tabular}{lcccccccccc}
\toprule
\multirow{2}{*}{\raisebox{-3pt}{\textbf{Method}}} & \multicolumn{2}{c}{\textbf{Coherence}} & \multicolumn{2}{c}{\textbf{Consistency}} & \multicolumn{2}{c}{\textbf{Fluency}} & \multicolumn{2}{c}{\textbf{Relevance}} & \multicolumn{2}{c}{\textbf{Average}} \\
\cmidrule(lr){2-3} \cmidrule(lr){4-5} \cmidrule(lr){6-7} \cmidrule(lr){8-9} \cmidrule(lr){10-11} 
 & $\rho$ & $\tau$ & $\rho$ & $\tau$ & $\rho$ & $\tau$ & $\rho$ & $\tau$ & $\rho$ & $\tau$ \\
\midrule
\multicolumn{11}{l}{\emph{Traditional Metrics}} \\
BLEU & 0.062 & 0.044 & 0.048 & 0.040 & 0.046 & 0.036 & 0.145 & 0.108 & 0.075 & 0.057 \\
ROUGE & 0.107 & 0.080 & 0.145 & 0.123 & 0.113 & 0.093 & 0.241 & 0.183 & 0.152 & 0.120 \\
BARTScore & 0.474 & 0.367 & 0.266 & 0.220 & 0.258 & 0.214 & 0.318 & 0.243 & 0.329 & 0.261 \\
BERTScore & 0.285 & 0.220 & 0.151 & 0.122 & 0.186 & 0.154 & 0.302 & 0.232 & 0.231 & 0.182 \\
BLEURT & 0.150 & 0.112 & 0.089 & 0.074 & 0.133 & 0.107 & 0.238 & 0.178 & 0.152 & 0.118 \\
CometKiwi & 0.353 & 0.273 & 0.151 & 0.124 & 0.207 & 0.170 & 0.203 & 0.151 & 0.228 & 0.180 \\
UniEval & 0.575 & 0.442 & 0.446 & 0.371 & 0.449 & 0.371 & 0.426 & 0.325 & 0.474 & 0.377 \\

\midrule
\multicolumn{11}{l}{\emph{Prompt LLM}} \\

G-Eval (GPT-3.5) & 0.440 & 0.335 & 0.386 & 0.318 & 0.424 & 0.347 & 0.385 & 0.293 & 0.409 & 0.323 \\
G-Eval (GPT-4) & 0.582 & 0.457 & 0.507 & 0.425 & 0.455 & 0.378 & 0.548 & 0.433 & 0.523 & 0.423 \\
GPT-3.5 & 0.459 & 0.371 & 0.393 & 0.331 & 0.355 & 0.296 & 0.455 & 0.363 & 0.415 & 0.340 \\
GPT-4 & 0.540 & 0.434 & 0.531 & 0.464 & 0.480 & 0.409 & 0.491 & 0.395 & 0.511 & 0.426 \\\vspace{3pt}
\makecell[l]{AUTOCALIBRATE\\~\citep{AUTOCALIBRATE}} & 0.570 & 0.493 & 0.500 & 0.467 & 0.487 & 0.452 & 0.560 & 0.483 & 0.529 & 0.474 \\ \vspace{3pt}
\makecell[l]{CoAScore$_{(n=10)}$\\~\citep{coascore}} & 0.541 & 0.419 & 0.339 & 0.299 & 0.367 & 0.308 & 0.478  & 0.379 & 0.431 & 0.351 \\ 
\makecell[l]{HD-EVAL-NN\\~\citep{HD-eval}} & 0.657 & - & 0.451 & - & 0.435 & - & 0.599 & - & 0.535 & - \\

\midrule
\multicolumn{11}{l}{\emph{Fine-tuned LLM}} \\
X-Eval & 0.530 & 0.382 & 0.428 & 0.340 & 0.461 & 0.365 & 0.500 & 0.361 & 0.480 & 0.362 \\
Prometheus & 0.150 & 0.126 & 0.150 & 0.137 & 0.189 & 0.168 & 0.164 & 0.138 & 0.163 & 0.142 \\
AUTO-J & 0.245 & 0.203 & 0.131 & 0.121 & 0.154 & 0.141 & 0.262 & 0.222 & 0.198 & 0.172 \\
TIGERScore & 0.381 & 0.318 & 0.427 & 0.387 & 0.363 & 0.327 & 0.366 & 0.304 & 0.384 & 0.334 \\
InstructScore& 0.328 & 0.276 & 0.232 & 0.213 & 0.260 & 0.237 & 0.211 & 0.179 & 0.258 & 0.226 \\
Themis (ours) & 0.566 & 0.485 & 0.600 & 0.566 & 0.571 & 0.533 & 0.474 & 0.412 & 0.553 & 0.499 \\

\bottomrule
\end{tabular}
\caption{Complete results on SummEval for Summarization task.}
\label{tab:Summarization}
\end{table*}

\begin{table*}
\centering
\small
\renewcommand{\arraystretch}{1.5}
\begin{tabular}{lcccccccccc}
\toprule
\multirow{2}{*}{\raisebox{-3pt}{\textbf{Method}}} & \multicolumn{2}{c}{\textbf{Context Maintenance}} & \multicolumn{2}{c}{\textbf{Interestingness}} & \multicolumn{2}{c}{\textbf{Knowledge Use}} & \multicolumn{2}{c}{\textbf{Naturalness}} & \multicolumn{2}{c}{\textbf{Average}} \\
\cmidrule(lr){2-3} \cmidrule(lr){4-5} \cmidrule(lr){6-7} \cmidrule(lr){8-9} \cmidrule(lr){10-11} 
 & $r$ & $\rho$ & $r$ & $\rho$ & $r$ & $\rho$ & $r$ & $\rho$ & $r$ & $\rho$ \\
\midrule
\multicolumn{11}{l}{\emph{Traditional Metrics}} \\

BLEU & 0.370 & 0.374 & 0.406 & 0.454 & 0.281 & 0.369 & 0.366 & 0.356 & 0.356 & 0.388 \\
ROUGE & 0.400 & 0.376 & 0.452 & 0.488 & 0.339 & 0.423 & 0.381 & 0.360 & 0.393 & 0.412 \\
BARTScore & 0.119 & 0.165 & 0.069 & 0.059 & 0.031 & 0.053 & 0.050 & 0.065 & 0.067 & 0.086 \\
BERTScore & 0.395 & 0.383 & 0.439 & 0.449 & 0.330 & 0.378 & 0.388 & 0.366 & 0.388 & 0.394 \\
BLEURT & 0.401 & 0.408 & 0.431 & 0.427 & 0.321 & 0.364 & 0.383 & 0.354 & 0.384 & 0.388 \\
comet22 & 0.491 & 0.496 & 0.544 & 0.544 & 0.407 & 0.450 & 0.489 & 0.492 & 0.483 & 0.496 \\
CometKiwi & 0.334 & 0.327 & 0.369 & 0.355 & 0.331 & 0.309 & 0.380 & 0.368 & 0.353 & 0.340 \\
UniEval & 0.595 & 0.613 & 0.557 & 0.605 & 0.536 & 0.575 & 0.444 & 0.514  & 0.533 & 0.577  \\

\midrule
\multicolumn{11}{l}{\emph{Prompt LLM}} \\
G-Eval (GPT-3.5) & 0.519 & 0.544 & 0.660 & 0.691 & 0.586 & 0.567 & 0.532 & 0.539 & 0.574 & 0.585 \\
G-Eval (GPT-4) & 0.594 & 0.605 & 0.627 & 0.631 & 0.531 & 0.551 & 0.549 & 0.565 & 0.575 & 0.588 \\
GPT-3.5 & 0.550 & 0.531 & 0.651 & 0.648 & 0.653 & 0.581 & 0.515 & 0.550 & 0.592 & 0.578 \\
GPT-4 & 0.680 & 0.680 & 0.822 & 0.779 & 0.810 & 0.786 & 0.769 & 0.739 & 0.770 & 0.746 \\
CoAScore$_{(n=20)}$ & 0.539 & 0.553 & 0.578 & 0.595 & - & - & 0.558 & 0.596 & - & - \\
HD-EVAL-NN  & 0.584 & 0.607 & 0.682 & 0.701 & 0.549 & 0.568 & 0.648 & 0.674 & 0.616 & 0.638\\

\midrule
\multicolumn{11}{l}{\emph{Fine-tuned LLM}} \\
X-Eval & 0.558 & 0.622 & 0.449 & 0.593 & 0.734 & 0.728 & 0.417 & 0.478 & 0.539 & 0.605 \\
Prometheus & 0.451 & 0.465 & 0.495 & 0.473 & 0.437 & 0.412 & 0.355 & 0.384 & 0.435 & 0.434 \\
AUTO-J & 0.452 & 0.449 & 0.490 & 0.459 & 0.339 & 0.357 & 0.425 & 0.437 & 0.427 & 0.425 \\
TIGERScore & 0.417 & 0.438 & 0.328 & 0.333 & 0.137 & 0.138 & 0.455 & 0.477 & 0.334 & 0.346 \\
InstructScore& 0.299 & 0.297 & 0.264 & 0.233 & 0.140 & 0.102 & 0.374 & 0.332 & 0.269 & 0.241 \\
Themis (ours) & 0.639 & 0.644 & 0.790 & 0.766 & 0.778 & 0.761 & 0.727 & 0.729 & 0.733 & 0.725 \\

\bottomrule
\end{tabular}
\caption{Complete results on Topical-Chat for Dialogue Response Generation task.}
\label{tab:Dialogue}
\end{table*}

\begin{table*}
\centering
\small
\renewcommand{\arraystretch}{1.5}
\begin{tabular}{lcccccccccc}
\toprule
\multirow{2}{*}{\raisebox{-3pt}{\textbf{Method}}} & \multicolumn{2}{c}{\textbf{SFHOT INF.}} & \multicolumn{2}{c}{\textbf{SFHOT NAT.}} & \multicolumn{2}{c}{\textbf{SFRES INFO.}} & \multicolumn{2}{c}{\textbf{SFRES NAT.}} & \multicolumn{2}{c}{\textbf{Average}} \\
\cmidrule(lr){2-3} \cmidrule(lr){4-5} \cmidrule(lr){6-7} \cmidrule(lr){8-9} \cmidrule(lr){10-11}
 & $\rho$ & $\tau$ & $\rho$ & $\tau$ & $\rho$ & $\tau$ & $\rho$ & $\tau$ & $\rho$ & $\tau$ \\
\midrule
\multicolumn{11}{l}{\emph{Traditional Metrics}} \\
BLEU & 0.070 & 0.054 & 0.055 & 0.040 & -0.023 & -0.018 & -0.004 & -0.004 & 0.024 & 0.018 \\
ROUGE & 0.107 & 0.082 & 0.075 & 0.055 & 0.118 & 0.090 & 0.105 & 0.078 & 0.101 & 0.076 \\
BARTScore & 0.211 & 0.162 & 0.130 & 0.094 & 0.265 & 0.201 & 0.226 & 0.165 & 0.208 & 0.156 \\
BERTScore & 0.135 & 0.104 & 0.126 & 0.093 & 0.157 & 0.120 & 0.139 & 0.102 & 0.139 & 0.105 \\
BLEURT & 0.219 & 0.171 & 0.229 & 0.171 & 0.244 & 0.186 & 0.282 & 0.211 & 0.244 & 0.184 \\
CometKiwi & 0.220 & 0.169 & 0.235 & 0.172 & 0.203 & 0.153 & 0.345 & 0.252 & 0.251 & 0.186 \\
UniEval & 0.249 & 0.191 & 0.320 & 0.238 & 0.225 & 0.169 & 0.333 & 0.247 & 0.282 & 0.211 \\

\midrule
\multicolumn{11}{l}{\emph{Prompt LLM}} \\
GPT-3.5 & 0.242 & 0.196 & 0.294 & 0.220 & 0.304 & 0.250 & 0.385 & 0.291 & 0.306 & 0.239 \\
GPT-4 & 0.302 & 0.263 & 0.359 & 0.283 & 0.213 & 0.178 & 0.405 & 0.316 & 0.320 & 0.260 \\
AUTOCALIBRATE & 0.357 & 0.313 & 0.440 & 0.383 & 0.315 & 0.272 & 0.416 & 0.351 & 0.382 & 0.330\\

\midrule
\multicolumn{11}{l}{\emph{Fine-tuned LLM}} \\
Prometheus & 0.169 & 0.141 & 0.211 & 0.171 & 0.161 & 0.134 & 0.150 & 0.122 & 0.173 & 0.142 \\
Auto-J & 0.176 & 0.152 & 0.127 & 0.106 & 0.179 & 0.153 & 0.084 & 0.070 & 0.141 & 0.120 \\
TIGERScore & 0.215 & 0.191 & 0.204 & 0.175 & 0.160 & 0.141 & 0.221 & 0.191 & 0.200 & 0.175 \\
InstructScore & 0.222 & 0.194 & 0.273 & 0.231 & 0.194 & 0.164 & 0.300 & 0.251 & 0.247 & 0.210 \\
Themis (ours) & 0.259 & 0.226 & 0.380 & 0.321 & 0.298 & 0.258 & 0.395 & 0.332 & 0.333 & 0.284 \\

\bottomrule
\end{tabular}
\caption{Complete results on SFRES \& SFHOT for Data to Text task.}
\label{tab:Data to Text}
\end{table*}

\begin{table*}
\centering
\small
\renewcommand{\arraystretch}{1.5}
\begin{tabular}{lccccccccc}
\toprule
\multirow{2}{*}{\raisebox{-3pt}{\textbf{Method}}} & \multicolumn{3}{c}{\textbf{CNN-DM}} & \multicolumn{3}{c}{\textbf{XSUM}} & \multicolumn{3}{c}{\textbf{Average}} \\
\cmidrule(lr){2-4} \cmidrule(lr){5-7} \cmidrule(lr){8-10} 
 & $r$ & $\rho$ & $\tau$ & $r$ & $\rho$ & $\tau$ & $r$ & $\rho$ & $\tau$ \\
\midrule
\multicolumn{10}{l}{\emph{Traditional Metrics}} \\
BARTScore & 0.732 & 0.680 & 0.555 & 0.175 & 0.171 & 0.139 & 0.454 & 0.425 & 0.347 \\
CometKiwi & 0.176 & 0.158 & 0.123 & 0.027 & 0.030 & 0.025 & 0.101 & 0.094 & 0.074 \\
UniEval & 0.682 & 0.662 & 0.532 & 0.461 & 0.488 & 0.399 & 0.572 & 0.575 & 0.466\\

\midrule
\multicolumn{10}{l}{\emph{Prompt LLM}} \\

G-Eval (GPT-4) & 0.477 & 0.516 & 0.410 & 0.211 & 0.406 & 0.343 & 0.344 & 0.461 & 0.377 \\
G-Eval (GPT-3.5) & 0.631 & 0.685 & 0.591 & 0.558 & 0.537 & 0.472 & 0.595 & 0.611 & 0.532 \\
GPT-3.5 & 0.454 & 0.514 & 0.417 & 0.279 & 0.348 & 0.295 & 0.366 & 0.431 & 0.356 \\
GPT-4 & 0.735 & 0.746 & 0.626 & 0.541 & 0.528 & 0.439 & 0.638 & 0.637 & 0.532 \\
AUTOCALIBRATE & 0.740 & 0.744 & 0.663 & 0.662 & 0.662 & 0.662 & 0.701 & 0.703 & 0.663\\

\midrule
\multicolumn{10}{l}{\emph{Fine-tuned LLM}} \\

Auto-J & 0.291 & 0.238 & 0.214 & 0.225 & 0.214 & 0.203 & 0.258 & 0.226 & 0.209 \\
TIGERScore & 0.574 & 0.562 & 0.479 & 0.424 & 0.445 & 0.412 & 0.499 & 0.504 & 0.446 \\
InstructScore& 0.287 & 0.278 & 0.233 & -0.096 & -0.134 & -0.119 & 0.095 & 0.072 & 0.057 \\
Themis (ours) & 0.747 & 0.761 & 0.680 & 0.599 & 0.607 & 0.546 & 0.673 & 0.684 & 0.613 \\

\bottomrule
\end{tabular}
\caption{Complete results on QAGS for Factuality Evaluation task.}
\label{tab:QAGS}
\end{table*}

\begin{table*}
\centering
\small
\renewcommand{\arraystretch}{1.5}
\begin{tabular}{lccccccccc}
\toprule
\multirow{2}{*}{\raisebox{-3pt}{\textbf{Method}}} & \multicolumn{3}{c}{\textbf{ROC}} & \multicolumn{3}{c}{\textbf{WP}} & \multicolumn{3}{c}{\textbf{Average}} \\
\cmidrule(lr){2-4} \cmidrule(lr){5-7} \cmidrule(lr){8-10} 
 & $r$ & $\rho$ & $\tau$ & $r$ & $\rho$ & $\tau$ & $r$ & $\rho$ & $\tau$ \\
\midrule
\multicolumn{10}{l}{\emph{Traditional Metrics}} \\
BLEU& 0.034 & 0.035 & 0.022 & 0.009 & 0.029 & 0.021 & 0.021 & 0.032 & 0.009 \\
ROUGE & 0.012 & 0.000 & 0.006 & 0.003 & -0.004 & -0.004 & 0.008 & -0.002 & 0.156 \\
BARTScore & 0.344 & 0.330 & 0.273 & 0.403 & 0.370 & 0.306 & 0.373 & 0.350 & 0.260 \\
BERTScore & 0.288 & 0.269 & 0.222 & 0.293 & 0.300 & 0.248 & 0.291 & 0.285 & 0.163 \\
BLEURT & 0.189 & 0.155 & 0.123 & 0.144 & 0.122 & 0.104 & 0.166 & 0.138 & 0.221 \\
CometKiwi & 0.246 & 0.218 & 0.176 & 0.289 & 0.283 & 0.238 & 0.268 & 0.251 & 0.176 \\

\midrule
\multicolumn{10}{l}{\emph{Prompt LLM}} \\
GPT-3.5 & 0.372 & 0.363 & 0.312 & 0.312 & 0.294 & 0.258 & 0.342 & 0.328 & 0.295 \\
GPT-4 & 0.590 & 0.578 & 0.518 & 0.382 & 0.368 & 0.336 & 0.486 & 0.473 & 0.260 \\
CoAScore$_{(n=20)}$ & - & - & - & - & - & - & 0.414 & 0.411 & 0.315 \\

\midrule
\multicolumn{10}{l}{\emph{Fine-tuned LLM}} \\

Prometheus & 0.031 & 0.013 & 0.014 & 0.001 & 0.001 & 0.003 & 0.016 & 0.007 & 0.146 \\
Auto-J & 0.460 & 0.454 & 0.410 & 0.308 & 0.306 & 0.278 & 0.384 & 0.380 & 0.284 \\
TIGERScore & 0.283 & 0.271 & 0.221 & 0.196 & 0.191 & 0.158 & 0.240 & 0.231 & 0.207 \\
InstructScore& 0.383 & 0.368 & 0.316 & 0.258 & 0.228 & 0.194 & 0.321 & 0.298 & 0.168 \\
Themis (ours) & 0.637 & 0.607 & 0.551 & 0.507 & 0.495 & 0.452 & 0.572 & 0.551 & 0.501 \\

\bottomrule
\end{tabular}
\caption{Complete results on MANS for Story Generation task.}
\label{tab:Story Generation}
\end{table*}

\begin{table*}
\centering
\small
\renewcommand{\arraystretch}{1.5}
\begin{tabular}{lccc}
\toprule
\multirow{2}{*}{\raisebox{-3pt}{\textbf{Method}}} & \multicolumn{3}{c}{\textbf{WMT23}} \\
\cmidrule(lr){2-4} 
 & $r$ & $\rho$ & $\tau$  \\
\midrule
\multicolumn{4}{l}{\emph{Traditional Metrics}} \\
BLEU& -0.130 & 0.021 & 0.018 \\
ROUGE & 0.081 & 0.151 & 0.117 \\
BARTScore & 0.091 & 0.118 & 0.093 \\
BERTScore & 0.123 & 0.219 & 0.170 \\
BLEURT & 0.163 & 0.263 & 0.208 \\
CometKiwi & 0.413 & 0.343 & 0.273 \\

\midrule
\multicolumn{4}{l}{\emph{Prompt LLM}} \\

GPT-3.5 & 0.388 & 0.347 & 0.278 \\
GPT-4 & 0.496 & 0.437 & 0.361 \\

\midrule
\multicolumn{4}{l}{\emph{Fine-tuned LLM}} \\

Prometheus & 0.144 & 0.129 & 0.107 \\
Auto-J & 0.128 & 0.104 & 0.087 \\
TIGERScore & 0.277 & 0.248 & 0.211 \\
InstructScore& 0.213 & 0.219 & 0.181 \\
Themis (ours) & 0.431 & 0.405 & 0.357 \\

\bottomrule
\end{tabular}
\caption{Complete results on WMT23 (zh-en) for Machine Translation task.}
\label{tab:Machine Translation}
\end{table*}

\end{document}